# Data-Driven Malaria Prevalence Prediction in Large Densely-Populated Urban Holoendemic sub-Saharan West Africa: Harnessing Machine Learning Approaches and 22-years of Prospectively Collected Data

Biobele J. Brown[1,2], Alexander A. Przybylski[4], Petru Manescu[4], Fabio Caccioli[4], Gbeminiyi Oyinloye[1,2], Muna Elmi[4], Michael J. Shaw[4], Vijay Pawar[4], Remy Claveau[4], John Shawe-Taylor[4], Mandayam A. Srinivasan[4], Nathaniel K. Afolabi[1,2], Adebola E. Orimadegun[1], Wasiu A. Ajetunmobi[1], Francis Akinkunmi[1], Olayinka Kowobari[1], Kikelomo Osinusi[1], Felix O. Akinbami[1], Samuel Omokhodion[1], Wuraola A. Shokunbi[3], Ikeoluwa Lagunju[1,2], Olugbemiro Sodeinde[1,2,4] and Delmiro Fernandez-Reyes[1,2,4,*]

1. Department of Paediatrics, College of Medicine University of Ibadan, University College Hospital, Ibadan, Nigeria.

2. Childhood Malaria Research Group, College of Medicine University of Ibadan, University College Hospital, Ibadan, Nigeria.

3. Department of Haematology, College of Medicine University of Ibadan, University College Hospital, Ibadan, Nigeria.

4. Department of Computer Science, Faculty of Engineering, University College London, Gower Street, London, WC1E 6BT, United Kingdom.

*Corresponding Senior Author: Delmiro.Fernandez-Reyes@ucl.ac.uk

**Short Title**. Data-Driven Locality-specific Elastic-Net based Monthly Malaria Prevalence Prediction.

**Key Words.** Digital Malaria Prevalence Prediction Systems; Monthly Malaria Prevalence Prediction; High-Transmission Holoendemic Malaria; Sub-Saharan West Africa; Supervised Machine Learning; Elastic Net Regression; Digital Global Health.

**ABSTRACT**

Plasmodium falciparum malaria still poses one of the greatest threats to human life with over 200 million cases globally leading to half-million deaths annually. Of these, 90% of cases and of the mortality occurs in sub-Saharan Africa, mostly among children. Although malaria prediction systems are central to the 2016-2030 malaria Global Technical Strategy, currently these are inadequate at capturing and estimating the burden of disease in highly endemic countries. We developed and validated a computational system that exploits the predictive power of current Machine Learning approaches on 22-years of prospective data from the high-transmission holoendemic malaria urban-densely-populated sub-Saharan West-Africa metropolis of Ibadan. Our dataset of >$9 \times 10^4$ screened study participants attending our clinical and community services from 1996 to 2017 contains monthly prevalence, temporal, environmental and host features. Our Locality-specific Elastic-Net based Malaria Prediction System (LEMPS) achieves good generalization performance, both in magnitude and direction of the prediction, when tasked to predict monthly prevalence on previously unseen validation data (MAE<=$6 \times 10^{-2}$, MSE<=$7 \times 10^{-3}$) within a range of (+0.1 to -0.05) error-tolerance which is relevant and usable for aiding decision-support in a holoendemic setting. LEMPS is well-suited for malaria prediction, where there are multiple features which are correlated with one another, and trading-off between regularization-strength L1-norm and L2-norm allows the system to retain stability. Data-driven systems are critical for regionally-adaptable surveillance, management of control strategies and resource allocation across stretched healthcare systems.

**INTRODUCTION**

Human malaria caused by *Plasmodium falciparum* is a mosquito-borne infectious disease threatening the lives of millions of people around the world. The World Health Organization (WHO) estimates that there were 212 million malaria cases globally in 2016, with 429,000 resulting in death. Of these, 90% of cases and 92% of deaths occurred in Africa, predominantly in sub-Saharan regions (with 76% and 75% of global cases and deaths occurring in only 13 countries) [1]. Around the world, children under 5 years-of-age are the most vulnerable, accounting for an estimated 70.6% of all malaria deaths in 2016 [2]. While various control and preventative interventions have been implemented over time, malaria still poses one of the greatest threats to human life.

One set of control measures includes surveillance and estimation of burden of disease to allow for strategic planning of already scanty healthcare and public health resources across endemic regions. Although the transformation of malaria surveillance into a core intervention has been designated as one of the three pillars of the Global Technical Strategy for malaria 2016-2030 (GTS) [3], current surveillance and predictive systems are inadequate at accurately capturing and estimating the extent of malaria, particularly in highly endemic countries [4].

There is an urgent need for predictive intelligent-systems that can reliably generalize future burden of malaria disease within well-defined *Plasmodium falciparum* malaria heavily-affected countries such as Nigeria in sub-Saharan West Africa. In Nigeria, the most populous country of Africa with 180 million inhabitants, the entire population is at risk of malaria (i.e. no malaria-free areas), with 76% of the population is deemed to be living in all-year-round high-transmission areas [5]. Nigeria is one of the most malaria burdened countries in the World accounting for 29% of Worldwide malaria cases and 26% of deaths in 2015 (mostly in children under five years of age), the largest proportion from any one country [1]. This global health challenge is particularly striking in large urban densely-populated cities such as Lagos (>15 million inhabitants) and Ibadan (>3 million inhabitants) both under large all-year-round malaria burden where stretched healthcare resources will benefit from advance knowledge of malaria burden (Figure 1).

Traditionally malaria-estimation systems have been centered around classical mathematical-models of disease dynamics with varying degrees of success. These models have been studied extensively and historically have provided the foundations of reasoning about and formalizing the dynamics of several infectious diseases. These have been pivotal to formulation of transmission models aimed at understanding relationships between the malaria parasite, the host and the vector. On the other hand, data-driven supervised machine learning algorithms fit models to a given dataset with the key aim of extrapolation or predicting the future based on past observations without the explicit incorporation of biological assumptions about the disease in question. This broad class of approaches are useful when explicit algorithmic solutions are not available, there is a lack of representative formal models, or the knowledge or concept about the application domain is poorly defined given its complexity such as in the case of malaria burden and disease dynamics. Machine Learning (ML) approaches, as opposed to the explicit mathematical-model driven ones, offer a well-established set of data-harnessing algorithms that are well-suited for capturing complex data patterns from which to perform generalizable predictive tasks.

A 2012 scoping review on systems for predicting malaria burden of disease [6] identified the use of mathematical modeling, regression, autoregressive integrated moving average and neural network approaches across 29 studies spanning 13 different countries, with varying populations, sample sizes and data sources. All these differed in key aspects such as input features, prediction models, model evaluation measures and their performance. More recent studies have explored machine learning methods alternative to neural networks, such as generalized linear models [7], random forests [8] and support vector machines [9–11] with varying degrees of success and using different and vastly heterogeneous datasets.

Currently, malaria surveillance predictive systems [4,6] as well as previous attempts of using machine learning approaches [7–11] have been severely hindered by the fact that collection of global data on malaria (vector, host, and environmental factors) is scanty, inaccurate and largely lacks quality control across all affected regions [4]. Only 10% of global malaria cases are reported through current surveillance systems [4]. This is also hindered by the challenges of access to reliable and accurate malaria diagnosis across malaria endemic regions. To tackle such challenges, we designed, developed and validated a malaria prevalence predictive system that exploits supervised machine learning approaches using a 22-year large quality-controlled and prospectively-collected malaria dataset that encapsulates a snapshot of the burden of *P. falciparum* malaria in the large densely-populated city of Ibadan, Nigeria.

Here we show that a data-driven ML framework is able to extract complex patterns among features of this large malaria burden snapshot to reliably predict next month malaria prevalence. We develop and validate a Locality-specific Elastic-Net based Malaria Prediction System (LEMPS) that shows good generalization performance, both in magnitude and direction of the prediction, when tasked to predict next month prevalence on previously unseen validation data. Regionally adaptable data-driven systems such as ours are key for aiding decision-support on planning and allocation of resources to local malaria clinical services and control strategies as well as assisting malaria burden surveillance. Our data-driven LEMPS simplicity and adaptability will facilitate its wide deployment to other settings in the sub-Saharan region to form a network of regionally-adaptable predictors.

## METHODS and MATERIALS

**Ethics Statement**

The internationally recognized ethical committee at the Institute for Advanced Medical Research and Training (IMRAT) of the College of Medicine, University of Ibadan (COMUI) approved this research on the platform of the Childhood Malaria Research Group (CMRG) within the academic Department of Pediatrics, University of Ibadan, as well as at school and Primary Care centers throughout the city of Ibadan with permit number: UI/EC/10/0130. Parents and/or guardians of study participants gave informed written consent in accordance with the World Medical Association ethical principles for research involving human subjects.

**Study Site**

The data used in this study has been routinely prospectively collected by the Department of Pediatrics of the College of Medicine of the University of Ibadan (COMUI), University College Hospital (UCH), Ibadan, Nigeria located in sub-Saharan West Africa (Figure 1a).

The city of Ibadan is a large urban metropolis with over three million inhabitants, the third largest city in Nigeria, with all-year-round (holoendemic) malaria transmission [12] (Figure 1b). Urban Ibadan is one of the most densely-populated areas in Nigeria. The city has a lengthy 8-month rainy season, with an average of 10 rainy days per month between May and October. Malaria transmission and clinical disease occurs throughout the year (Figure 1b).

UCH-Ibadan is the largest and main academic hospital in urban Ibadan as well as the first teaching hospital in Nigeria. It is a basin-of-attraction for healthcare from all regions of this large urban setting of Ibadan. It is easily and readily accessible using largely available public transport, and the cost and standards of care make it the primary choice by large sectors of the population. UCH-Ibadan is an all-clinical-services 800-bed tertiary-care hospital that also provides all-specialties secondary-care as well as some primary-and-community care services (Figure 1a).

**Study Design**

We routinely screen for malaria and parasite-density using Giemsa blood thick and thin films all children up to 16 years-of-age attending to any of our well-children or ill-children services. Our clinical services are: emergency ward; in-patient wards; out-patient clinics; routine school well-children malaria screening activities as well as secondary and primary care screening. Every year, we carry-out an approximate $5x10^3$ malaria microscopy screens across all our clinical services listed above. The data used in this study includes all those screened in all our services from January 1996 to December 2017 inclusive, a total of 22 years (Tables 1, 2 and 3).

**Dataset Characteristics**

Demographics (year, month and age) and malaria clinical data (malaria diagnosis and parasite density) used in this study have been continuously collected between January 1996 and December 2017 as explained in previous section. Overall demographic yearly aggregates are given in Table 1.

For this study, our Ibadan dataset was processed to consists of the monthly aggregated variables from larger datasets collected under our standardized routine malaria-screening which is linked to our clinical care pathways and departmental surveillance figures [12–18] (Tables 1, 2 and 3). Our prospectively collected dataset is linked to and amalgamates our childhood malaria case-control and longitudinal studies and bio-banks [12–19], as well as our research and development of an fast automated machine-learning-driven optical-malaria-diagnostic microscope [20]. The aggregated data used in this study is described in Tables 1, 2 and 3.

We assembled our full Ibadan dataset, denoted by *D*, by aggregating data for each month from January 1996 to December 2017 (22 years), creating thus a total of 264 (22x12) entries (Tables 1 and 2), each containing the following 15 variables (Table 3) namely: 1) year (not aggregated); 2) month (not aggregated); 3) total number screened (sum); 4) median age (months) of malaria-negative; 5) median age (months) of malaria-positive; 6) age (months) inter-quartile range of malaria-negative; 7) age (months) inter-quartile range of malaria-positive; 8) mean blood parasite density (MPs/µl); 9) standard deviation of blood parasite density (MPs/µl); 10) total rainfall (mm); 11) proportion of that year total rainfall; 12) minimum temperature (C°); 13) maximum temperature (C°); 14) mean temperature (C°) and 15) malaria prevalence (proportion of those who were screened and have confirmed malaria).

Our full Ibadan dataset is therefore a matrix *D* where each entry or row-instance of *D* is represented by a 1 by *N*=15 vector *d* encoding the variables [*year*, *month*, *number-screened*, *median-age-neg*, *median-age-pos*, *iqr-age-neg*, *iqr-age-pos*, *x-pd*, *sd-pd*, *mm-rf*, *mmP-rf*, *min-temp*, *max-temp*, *x-temp*, *prep*] (Tables 1, 2 and 3).

**Malaria Screening**

Malaria parasites (MPs) were detected and counted by microscopy following Giemsa staining of thick and thin blood films [12–18,21]. The criterion for declaring a participant to be malaria parasite-free was no detectable parasites in 100 high-power (100x) fields in both thick and thin films. We validated the diagnosis outcome by randomly selecting one in ten thick blood films for independent review by local external experienced senior malaria-microscopy technologists. Parasite Density (PD), malaria parasites per microliter (MPs/µl), are calculated by dividing the number-of-observed MPs by the number-of-counted White Blood Cells (WBC) and then multiplied by 8x10$^3$ as per widely established [12–18,21].

**Environmental Variables**

All of Ibadan's weather variables (rainfall, temperature) have been acquired from the International Institute for Tropical Agriculture (IITA) Ibadan, Nigeria; (web: www.iita.org) that keeps Ibadan's records since 1967.

**Dataset Features and Encoding of Prediction Tasks for Supervised Machine Learning**

The full Ibadan dataset *D* comprises of the two following datasets: 1) A Training Set (DTRAS) containing all the instances from the years 1996 to 2014 (19 years) as a *M* by *N* matrix where *M* = 19 x 12 = 228 row-vector instances and *N* = 15 variables (Tables 2 and 3) and; 2) A Validation Set (DVALS) containing all the instances from the years 2015 to 2017 (3 years) as a *M* by *N* matrix where *M* = 3 x 12 = 36 row-vector instances and *N* = 15 (Tables 2 and 3).

Supervised trained ML models aim to predict a target outcome from a set of new unseen input variables (commonly referred to as features). The goal is to train a predictive system that, given the aggregated variables $\hat{X}$ collected at the end of the Month M1, can reliably estimate the malaria prevalence of the following Month (Task+1) $\hat{y}$ which is our target outcome. We therefore encoded the datasets DTRAS and DVALS as a supervised learning regression task. The task is denoted by the encoded Dataset Task+1 M1 (DT1M1) (Table 4) such that the relationship between the features $X$, a row-vector containing the aggregated variables for one month, and the outputs $y$, the prevalence observed the following month, can be learned by training a supervised ML model (Tables 3 and 4).

DT1M1-DTRAS (Table 4) is a *M*=228 by *N*=16 matrix where each row-vector instance $d_{i,j} = DTRAS(X_{i,j}, y_{i,j+1})$ where *i*=year (from 1996 to 2014), *j*=month (January to December), $X_{i,j}$ is a 1 by *N*=15 row-vector containing all the variables for that month *j* of the year *i* (including the prevalence value of that unique month) and $y_{i+,j+}$, with $\begin{cases} i_+ = i + (j+1)/12 \\ j_+ = (j+1)/12 \end{cases}$, is the malaria prevalence target value of the following next month. Similarly, DT1M1-VALS (Table 4) is a *M*=36 by *N*=16 matrix where each row-vector instance $d_{i,j} = DVALS(X_{i,j}, y_{i+,j+})$ where *i*=year (2015 to 2017), *j*=month (January to December), $X_{i,j}$ is a 1 by *N*=15 row-vector containing all the variables for that unique year and month and $y_{i+,j+}$ is the malaria prevalence target value of the following unique month.

To exploit information from previous months, we encoded six regression tasks M*m* (*m* = 1 to 6) that, while still predicting $y_{i+,j+}$ malaria prevalence of the following *i*=year *j*=month, add to that unique (*i, j*) month instance the 1 by *N* (N = number-of-variables = 15) row-instance variables of the previous (*m*-1) months where *m* = 1 to 6 (Table 4). The resulting dimensionality of (*i, j*) instance row-vectors for each M*m* task are 1 by *N* = [(number-of-variables x *m*) + 1] = [(15 x *m*) + 1] (Table 4). This encoding creates the following tasks Tasks+1 M1 to M6 (DT1M1, DT1M2, DT1M3, DT1M4, DT1M5 and DT1M6) for both DTRAS and DVALS datasets (Table 4). Finally, the DTRAS and DVALS Task+1 datasets were further processed to remove the year (non-aggregated) column(s) variable (Table 4).

**Supervised Machine Learning Regression Approaches.**

To build the predictive regression system we used Generalized Linear Models (GLM), Ensemble Methods (EM) and Support Vector Machines (SVM) within a supervised learning framework (Figures 2 and 4).

**Generalized Linear Models (GLM):**

We used a set of GLM regression algorithms in which the target outcome $y$ (dependent variable) is expected to be a linear combination of the input features (independent variables). A GLM such as linear regression (LR) or ordinary least squares (OLS) fits a linear model with vector coefficients $w = (w_1, ..., w_p)$, where $p$=number-of-dimensions, that minimizes the residual sum of squares between target outcome $y$ and the predicted outcome $\hat{y}$ of the lineal model $\min_{w} \| Xw - y \|_2^2$ where $X$ is a $M$ by $p$ matrix with $X_i$=1 to $M$ row-vectors containing the features $p$ of $i^{th}$-example and $y$ is a $M$ by 1 column-vector containing $y_i$ target value of the $i^{th}$-example.

*Ridge Regression (RR)*: RR imposes a L2-norm penalty on the regression coefficients. The ridge coefficients minimize a penalized sum of squares $\min_{w} \| Xw - y \|_2^2 + \alpha \| w \|_2^2$ where $\alpha \geq 0$ controls the amount of shrinkage.

*Least Absolute Shrinkage and Selection Operator (LASSO)*: The LASSO estimate solves the minimization of the least-squares penalty with a L1-norm prior regularization added. The optimization problem is $\min_{w} \dfrac{1}{2n_{samples}} \| Xw - y \|_2^2 + \alpha \| w \|_1$ where $\alpha$ is a constant that controls the degree of sparsity of the coefficients estimated and $\| w \|_1$ is the L1-norm of the parameter vector.

*Elastic Net (EN)*: The EN takes advantage of both RR (L2-norm) and LASSO (L1-norm) prior regularization. The combination has the capacity to learn sparse models with few non-zero weights such as in LASSO while maintaining regularization properties of RR. The convex combination of L1-norm and L2-norm is controlled by the L1-ratio parameter $r$. The optimization problem is:

$$\min_w \frac{1}{2n_{samples}} \| Xw - y \|_2^2 + \alpha\rho \| w \|_1 + \frac{\alpha(1-\rho)}{2} \| w \|_2^2$$

The EN has useful properties where there are multiple features correlated with one another. While LASSO will pick one of these at random, the EN is more likely to pick the correlated features.

*Least Angle Regression (LARS)*: Similar to forward stepwise regression, at each step it finds the predictor most correlated with the response, but it continues in a direction that is equiangular between predictors when there are multiple equally correlated predictors [22].

*LARS-LASSO*: Implements LASSO using the LARS algorithm [22] instead of a coordinate descent.

**Ensemble Methods**

*Random Forests (RF)*:

RF is a type of bagging ensemble method whereby a set of decision trees are learned on randomly sampled subsets of features and training data points. As in bootstrap sampling, data samples are drawn with replacement, helping to ensure the base learners are de-correlated. This method aims to reduce variance of the estimator and decrease issues of over-fitting common to standard decision trees. Predictions are then made by averaging the predictions from the set of individual trees [23]. The number of trees in the forest can be tuned as a parameter, as well as parameters of the individual trees themselves. These include 1) the maximum number of features considered when splitting at a node; 2) the maximal depth a tree can be grown until; 3) the minimum number of data samples necessary for a split; 4) the minimum number of data samples at leaf nodes.

**Support Vector Machines**

*Support Vector Regression (SVR)*: belongs to the family of Support Vector Machine learning algorithms [24], adapted for predicting continuous outputs[25]. The problem can be described as fitting a function $f(x)$ to a set of training examples such that any deviations from the observed targets are less than $\varepsilon$ which defines a margin around predictions where deviations are not penalized. The aim is to maximize the margin for a lower generalization error. The final solution results in a subset of training points that define the margin, the so-called support vectors.

Considering a linear function $f(x) = \langle w, x \rangle + b$, the norm $\|w\|^2$ can be minimized

$$\frac{1}{2}\|w\|^2 + C\sum_{i=1}^{l}(\xi^+ + \xi^-) \text{ subject to } \begin{cases} y_i - \langle w, x \rangle - b \leq \varepsilon + \xi^+ \\ \langle w, x \rangle + b - y_i \leq \varepsilon + \xi^- \\ \xi^+, \xi^- \geq 0 \end{cases}$$

where slack variables $\xi^+, \xi^-$ are incorporated to allow for errors outside the $\varepsilon$ range and form a constrained optimization problem. A constant $C \geq 0$ acts as a regularization parameter, controlling the bias-variance trade-off or the model fit.

We used a radial basis function kernel $K(x, x') \exp(-\gamma \|x - x'\|^2)$ for the implicit mapping of features onto high or infinite dimensions where the parameter $\gamma > 0$ is the Gaussian and thus can be used to control how much data points influence the function fit.

The optimal solution is $\hat{w} = \sum_{i=1}^{N} a_i x_i$ so that future predictions can be made by

$$f(x) = \hat{b} + \sum_{i,j=1}^{N} a_i K(x_i, x_j)$$

**Error Measures and Parameter Tuning**

Mean Absolute Error (MAE) and Mean Square Error (MSE) measures were used when evaluating the quality of predictions of malaria prevalence.

MAE measures the average magnitude of the errors in a set of predictions without considering their direction. MAE is a risk metric corresponding to the expected value of the absolute error loss or L1-norm loss:

$$MAE(y_{i,j}, \hat{y}_{i,j}) = \frac{1}{N_y} \sum_{i=1}^{N_y} \frac{1}{12} \sum_{j=1}^{12} |y_{i,j} - \hat{y}_{i,j}|$$

where $N_y$ is the number-of-years used for evaluation, $y_{i,j}$ is the true prevalence value of the month *j of the year i* and $\hat{y}_{i,j}$ is the corresponding predicted prevalence value. MAE is a scale dependent measure which is intuitive to interpret and its inclusion in malaria forecasting evaluation has been recommended [6].

MSE measures the average of the squares of the errors in a set of predictions incorporating both variance on the estimator and its bias. MSE is a risk metric corresponding to the expected value of the squared (quadratic) error or loss:

$$MSE(y_{i,j}, \hat{y}_{i,j}) = \frac{1}{N_y} \sum_{i=1}^{N_y} \frac{1}{12} \sum_{j=1}^{12} (y_{i,j} - \hat{y}_{i,j})^2$$

where *N* is the number-of-years used for evaluation, $y_{i,j}$ is the true prevalence value of the month *j of the year i* and $\hat{y}_{i,j}$ is the corresponding predicted prevalence value. MSE is also a scale dependent measure and is a commonly used indicator of trained model performance within a study.

For LASSO-LARS alpha parametrization we also used Akaike Information Criterion (AIC) [26] and the Bayesian Information Criterion (BIC) [27] both used for model selection.

**Algorithm Parametrization, Evaluation and Model Selection.**

An overview of the training framework for algorithm parametrization and selection is shown in Figure 2. For algorithm parametrization and evaluation, each of the training DTRAS datasets encoding the DT1M1 to M6 regression tasks, DT1M1-DTRAS to DT1M6-DTRAS (Table 4), were randomly split $10^3$ times into a Train Set (TS) containing 75% of the instances and a Held-Out Test Set (HOTest) containing 25% of the instances (Figure 2). The TS is a $M$ by $N$ matrix where $M$ = ceiling (0.75 x number-of-instances) and $N$ = (number-of-variables per each T1M1 to M6 tasks) Table 4. The HOTest is a $M$ by $N$ matrix where $M$ = (rest of the number-of-instances not in TS) and $N$ = (number-of-variables per each T1M1 to M6 tasks).

Each TS [X, y] was then used for the parameterization of each regression task algorithm within the framework (Figure 2). For tuning the hyper-parameter alpha (regularization strength) for RR, a set of alphas = [$1^{-3}$, $1^{-2}$, $1^{-1}$, 1, 10, $10^2$] were used and the best alpha selected by 5-fold cross-validation on the TS (Figure 2). For alpha selection in LASSO, EN, LARS, LASSO-LARS, we used model-specific iterative fitting along regularization path and selecting the best model by 5-fold cross-validation on the TS (Figure 2). Selection of best parameters was carried out using MSE as implemented in the scikit-learn python library [28].

For LASSO-LARS alpha parametrization we also used Akaike Information Criterion (AIC) and the Bayesian Information Criterion (BIC) (Figure 2). These provide an optimal estimate of the regularization parameter by computing a single regularization path instead of several when k-fold cross-validation was used. The AIC and BIC criteria are useful for selecting the value of the regularization parameter by making a trade-off between the goodness of fit and the complexity of the model.

We parametrized the meta-estimator RF with number-of-trees=10; maximum-features = number-of-features; nodes are expanded until all leaves are pure or until all leaves contain less than 2; using bootstrap when building trees (Figure 2).

For SVR we used a gaussian kernel and carried out 5-fold cross-validation to parametrize C and $\gamma$ with the following grid search C = [1, 10, $10^2$, $10^3$, $10^4$] and $\gamma$ = [1, $10^{-1}$, $10^{-2}$, $10^{-3}$, $10^{-4}$] respectively.

After each parametrization, the algorithm was trained on the TS with the optimal parameters and predictions were made on the target outcome $\hat{y}$ (prevalence of following month) on the X instances of HOTest (Figure 2). The trained algorithm test performance was then measured by MAE and MSE (Figure 2) and mean$\pm$SD of MAE and MSE over the $10^3$ random splits of DT1M1-DTRAS to DT1M6-DTRAS (Figure 3).

**L1-L2 Ratio and regularization strength Elastic Net Parametrization**

After selecting EN as the main ML algorithm for the system, we parametrized both *a* (regularization strength) and the L1-norm to L2-norm ratio (L1Ratio) as illustrated in Figure 4a. Each of the training DTRAS datasets encoding the DT1M1 to M6 regression tasks, DT1M1-DTRAS to DT1M6-DTRAS, were randomly split $10^3$ times into a Train Set (TS) containing 75% of the instances and a Held-Out Test Set (HOTest) containing 25% of the instances (Figure 4a). The TS is a *M* by *N* matrix where *M* = ceiling (0.75 x number-of-instances) and *N* = (number-of-variables per each T1M1 to M6 tasks). The HOTest is a *M* by *N* matrix where *M* = (rest of the number-of-instances not in TS) and *N* = (number-of-variables per each T1M1 to M6 tasks).

For EN $a$ and L1Ratio we used model-specific iterative fitting along regularization path and selecting the best model by 5-fold cross-validation on the TS (Figure 4a). Selection of best parameters was carried out using MSE as implemented in the scikit-learn python library [28]. After each parametrization, the EN was trained on the TS with the best parameters and predictions for target outcome $\hat{y}$ (prevalence of following month) were made on the X instances of HOTest (Figure 4a). The trained algorithm test performance was then measured by MAE and MSE (Figure 4a). The mean±SD of MAE, mean±SD of MSE, mean±SD of $a$ s and median±IQR of L1Ratio were plotted over the $10^3$ random splits of DT1M1-DTRAS to DT1M6-DTRAS (Figure 5). The $y_i$ (true prevalence value of all instances $i$) versus the mean of $\hat{y}_i$ (mean predicted prevalence value over the times the instance $i$ was included in the HOTest) is plotted in Figure 6 for all regression tasks.

**Validation of the Elastic Net Framework**

We show the validation of the trained EN on the regression tasks DT1M1-DTRAS to DT1M6-DTRAS datasets using respective validation datasets DT1M1-DVALS to DT1M6-DVALS (Figure 4b, 7, 8 and 9) by using the task specific $a$ and L1Ratio parameters (Figure 5c and 5d).

**Software and Libraries**

We coded the framework using the Python programming language version 3.6.5 and with the open-source scikit-learn (version 0.19.1) machine-learning library [29]. Anaconda Navigator 1.8.4 was used to standardize python environment (Python 3.6.5, Anaconda, Mar 29[th], 2018). Dataset aggregates and statistical analysis were carried out with StataSE v12 and Python scripts; plots and figures were plotted with GraphPad and Miner3D. Software available upon request.

**RESULTS**

**Study Participants**

By orchestrating our wide clinical and community malaria screening services with our data collection protocols and malaria diagnosis quality standards, we have ensembled a large and fine-grained dataset that encapsulates the burden of malaria disease within an urban densely-populated all-year-round high-malaria-transmission setting, the city of Ibadan, in the sub-Saharan West African region[12] (Figure 1a and 1b). The city of Ibadan is the third largest city in Nigeria, with over three million inhabitants. The city experiences a lengthy 8-month rainy season, with an average of 10 rainy days per month between May and October where malaria transmission and clinical disease occurs throughout the year (Figure 1b).

In urban high-transmission holoendemic settings such as Ibadan, the burden of malaria vastly falls on children (Table 1). Although malaria predominantly affects children under 5 years-of-age, there is also large burden in children up to 16 years-of-age. Therefore, we routinely screen for malaria all children attending to any of our well-children or ill-children services across the city of Ibadan (Figure 1a). The data used in this study includes those screened in our services from January 1996 to December 2017 inclusive, a total of 22 years (Figure 1b and Table 1). The Ibadan 22-years dataset used in this study is supported by the screening of >$9 \times 10^4$ study participants (Table 1, Figure 1b). Overall yearly aggregates of clinical demographics are described in Table 1. The Ibadan dataset $D$ consists of a training set DTRAS with >$8 \times 10^4$ subjects and a validation set DVALS with >$1 \times 10^3$ subjects (Tables 1 and 2). Figure 1b shows that despite Ibadan's malaria burden has decreased over the last 22 years. However, the city of >3 million inhabitants (predominantly children) is still under a significantly large all-year-round burden of the disease, currently >5% at its lowest during dry-season months of December and January.

In the following results sections, we demonstrate a data-driven supervised machine learning system that is suitable to learn from the Ibadan dataset to provide predictions of the burden of the disease by estimating the prevalence of the month ahead.

**The Elastic Net consistently estimates with low error next-month prevalence across all regression tasks on training dataset.**

To select a supervised machine learning approach suitable for the task of predicting the malaria prevalence of the next month, we parametrized and trained nine algorithms [EN, LASSO, RR, LARS, LASSO-LARS, LASSO-LARS-AIC, LASSO-LARS-BIC, RF, SVR] (Figure 2) on six regression tasks DT1M1 to DT1M6 (Table 4) using the DTRAS dataset (DT1M1-DTRAS to DT1M6-DTRAS) carrying out held-out test over $10^3$ random splits of the datasets (Figure 2). The mean±SD of MAE and mean±SD MSE for each algorithm and for each regression task is shown in Figure 3.

EN (Figure 3, filled circles); LASSO (Figure 3, filled squares); LASSO-LARS (Figure 3, filled rhomboid) and LASSO-LARS-AIC (Figure 3, empty circles), predictors performed consistently with low MAE ($\leq 6.1 \times 10^{-2}$) and low MSE ($\leq 6.8 \times 10^{-3}$) across all the regression tasks. RR predictors (Figure 3, filled up-triangles) slightly decreased performance at regression tasks DT1M3 to DT1M6. LARS predictors (Figure 3, filled down-triangles) were worst at the largest dimensionality of the task DT1M6. LASSO-LARS predictors using the BIC information measure (Figure 3, empty squares) consistently performed worse than LASSO-LARS-AIC. RF predictors (Figure 3, empty up-triangle), despite being consistent across tasks, performed slightly worse than EN, LASSO and RR. SVR predictors (Figure 3, empty down-triangle) were also consistent across all regression tasks, but significantly had the worst performance when compared to all other algorithms on each of the tasks DT1M1 to DT1M2 with MAE ($1.1 \times 10^{-1}$) and MSE ($1.8 \times 10^{-}$

²). For tasks DT1M1 and DT1M2, all predictors (except SVR) performed with MAE (£6.8x10⁻²) and MSE (£8x10⁻³) Figure 3.

**Building a Locality-specific Elastic-Net based Malaria Prediction System (LEMPS)**

Using our Ibadan DTRAS dataset, we show that EN regularization-strength and L1-norm parametrization produce next-month prevalence estimates with low error and allows us to build a regionally adaptable Locality-specific EN based Malaria Prediction System (LEMPS).

We have chosen to concentrate our predictive model LEMPS on the EN algorithm firstly because: 1) EN achieved consistently good performance across all DT1M1 to DT1M6 (Figure 3, Figure 5) in the DTRAS data with MAE (£6.1x10⁻²) and MSE (£6.8x10⁻³) and 2) the L1-norm Ratio (that controls L1-norm vs L2-norm regularization) could be indeed useful in fine-tuning the system as the dynamics of the burden of disease change and/or it is used in another locality. For building LEMPS, the EN was parametrized for [$a$, L1Ratio] on the six regression tasks DT1M1 to DT1M6 using the DTRAS datasets (DT1M1-DTRAS to DT1M6-DTRAS) and carrying out held-out test over 10³ random splits of the datasets (Figure 4a). The mean±SD of MAE, mean±SD MSE, mean±SD of $a$ and median±IQR of L1Ratio for each regression task are shown in Figure 5. L1Ratio tuned EN achieved similar consistent performance to those shown in the previous section with MAE £6x10⁻² and MSE £6.5x10⁻³ (Figures 3, 4a, 5a and 5b). For each task DT1M1 to DT1M6 the performance was achieved by a unique range of $a$ and L1Ratios (Figures 4a, 5c and 5d) and these parameters were used for building and validating the final LEMPS system as described in the next section.

For each regression task DT1M1 to DT1M6 the $y$ (true prevalence value) and the mean $\hat{y}$ (predicted prevalence value) over 10³ random splits of DTRAS was plotted (Figure 6a to 6f dotted-black line and red line respectively). Over 10³ random splits of DTRAS each instance fell into the HOTest between 225 to 285 times. The mean of an instance predicted prevalence value, over the times that instance fell into the HOTest shows an overall good alignment with the true prevalence value of such instance across all tasks (Figure 6a to 6f). This is consistent with the low MAE and MSE values observed (Figure 5a and 5b). However, there are small subsets of instances that carry most of the error as follows: 1) Labeled as (1) in Figure 6a to 6f, from the 1996 to the 2000 period rain-season where the trained LEMPS, despite agreeing with the direction, underpredicted prevalence; 2) Labelled as (2) in Figure 6a to 6f, during 1996 period dry-season, despite agreeing with the direction, the trained LEMPS underpredicted prevalence and; 3) Labelled as (3) in Figure 6a and 6b (only DT1M1 and DT1M2), during the 2011 dry-season period, the trained LEMPS did not agree with the direction and overpredicted prevalence.

**Validating the Locality-specific Elastic-Net based Malaria Prediction System (LEMPS)**

The Elastic Net based system trained on best hyper-parameters estimates next-month prevalence with low error across all regression tasks on the 2015, 2016 and 2017 validation datasets. For each task DT1M1 to DT1M6 the performance was achieved by a unique range of $a$ and L1Ratios (Figure 5c and 5d), information that was then used for building and validating the final system (Figure 4b). The LEMPS mean of $a$ and the median of L1Ratio values obtained in the previous section (Figure 4a, Figure 5c and 5d) were chosen to build and validate the final LEMPS system on a previously unseen set of instances from the 2015, 2016 and 2017 period, the DVALS dataset, as shown in Figure 4b. For each DT1M1 to DT1M6 task an EN was trained using the full DTRAS dataset with selected parameters and its monthly performance was assessed on the 2015, 2016 and 2017 DVALS (Figures 7, 8 and 9). On all regression tasks, the LEMPS monthly prevalence predictions achieved consistently low MAE ($\leq 6\times10^{-2}$), low MSE ($\leq 7\times10^{-3}$) with Pearson Correlation Coefficient (PCC) ranging between 0.4 to 0.8 (Figure 7).

To assess the quality and direction of these monthly validation predictions, a scatter 2D plot of predicted prevalence value versus true prevalence value for all DT1M1 to DT1M6 prediction tasks is shown in Figure 8 where red and blue dots represent rainy and dry season months respectively. The plots highlight the importance of interpreting the validation of the predictions in relation to the problem domain. For example, validation year 2017 shows very good prediction agreement (i.e. dots closer to the diagonal) except for two months (one rain season and one dry season) which impairs its overall yearly PCC (Figure 7). Therefore, to further evaluate these predictions within an error-tolerance which is relevant for making the system suited and usable for a high-transmission holoendemic setting, we plotted the predicted monthly prevalence for all tasks against the true prevalence with a +0.1 to -0.05 tolerance error (Figure 9). Overall, across all 216 monthly predictions on the 2015 to 2017 validation set (3 years x 12 months x 6 tasks), 80% were within the tolerance error +0.1 to -0.05 (Figure 9) which is relevant for this holoendemic region and makes our system extremely usable for decision support in the Ibadan setting. During the long Ibadan rainy season (April to November), the LEMPS is extremely robust (95% of predictions within range) in estimating monthly prevalence within the error-tolerance range (Figure 9), except during the month of September 2017 where extreme prediction outliers (Figure 9**) made us suspect a critical event. We discovered that during that month, a country-wide general Nigerian Federal Government healthcare system strike had a nation-wide effect on our clinics. This reinforces the usefulness of our proposed system as a novelty detection system as in years 2015 and 2016 the LEMPS was robust in estimating September's month prevalence (Figure 9). During the dry season (December to March) the system also performs consistently within the error tolerance boundaries during the months December to February. However, during the month of March, for all validation years some prediction tasks underpredicted below the -0.05 range, an effect that is most extreme on 2017 prediction (Figure 9*). The month of March is the transition boundary from the dry to the rainy season and despite the trained LEMPS mostly agreeing with the direction of the prediction, the magnitude of the estimates for year 2017 were on the -0.1 range instead of -0.05. We could not find a critical event explanation for such observation.

**Use Cases and Deployment Analysis of Locality-specific Elastic-Net based Malaria Prediction System (LEMPS)**

The LEMPS system is easily deployable in current off-the-shelf hardware and opens the door for sustainable digital global health. The system could be further trained, deployed and developed using free open-source Python and ML tools provided within the freely available Anaconda Navigator environment. We propose a use case where each regional health center is a locally trained EN node (harnessing such local data at its best) within an interconnected network of EN predictors, via a distributed ledger, where new nodes could use closer regional predictors while they refine their own predictors (Figure 10).

The simplicity of LEMPS provides an incentive for sub-Saharan centers by giving decision-support value to their own routinely collected malaria data. This in turn should encourage those centers to transfer such data (15 variables in this study) into simple digital format that can be exploited by themselves and by the network of LEMPS predictors (Figure 10). As the network of locally specialized LEMPS predictors grows, it opens the possibility of meta-learning and novelty detection algorithms to be applied for tasks such as early epidemic prediction and more efficient distribution of resources across malaria affecting regions.

**DISCUSSION**

We have designed, developed and validated a machine-learning based system that is able to reliably predict next month malaria prevalence within urban densely-populated holoendemic malaria Ibadan with low error. The Locality-specific Elastic Net based Malaria Prediction System (LEMPS) shows good generalization performance, both in magnitude and direction of the prediction, when tasked to predict monthly prevalence of previously unseen data from years 2015, 2016 and 2017.

Our system exploits 19 years [1996 to 2014] of host information (age, malaria status, parasite densities); temporal information (year, month) and; environmental information (rainfall, temperature), from a predominantly Yoruba, largely-populated well-defined spatial urban setting living under high all-year-round malaria burden. We used our locality-specific data, the Ibadan dataset, to train a relevant LEMPS which currently contributes to decision making on managing our site malaria healthcare and surveillance resources. The trained LEMPS has an error-tolerance within +0.1 to -0.05 across all prediction tasks which is appropriate for a system to be usable in the high-transmission holoendemic setting of Ibadan. During the long Ibadan rainy season (April to November) the system is extremely robust in estimating monthly prevalence within the error-tolerance range. The system has also shown novelty-detection capabilities by highlighting prediction outliers observed in collection of validation data from September 2017 which was affected by a personnel strike in the healthcare system. Interestingly, the system has shown the complexity of the dynamics of the burden of disease near the dry-to-rainy season transition period (i.e. March). This may be due to emerging patterns across this seasonal transition period as we have observed recent dramatic changes of environmental factors in the city of Ibadan. Furthermore, recent investment on Ibadan's infrastructure may be playing a role in these changes. We expect that feature enrichment refinements focused on transition periods will allow the system to further improve its accuracy. These adjustments will have to take into account that, despite Ibadan's malaria burden has decreased over the last 22 years, the city is still under a currently changing but still significantly-large all-year-round burden of the disease (Figure 1b).

We have shown that a data-driven machine learning approach offers an alternative that allows for the creation of predictive systems that do not rely on an explicit formulation of the disease process. We focused our system on the Elastic Net, as it produced stable results across all prediction tasks while also providing flexibility of tuning regularization strength as well as the L1- to L2-norm ratio. The EN is well suited for problems such as malaria prediction, where there are multiple features which are correlated with one another, and trading-off between L1-norm (LASSO) and L2-norm (RR) allows the system to retain stability. We show that the EN based system provides an efficient, yet flexible, system for all the regression tasks relevant to the clinical and epidemiological context within the region.

Previous ML systems [7–11] have used significantly smaller datasets and none have harnessed features such as host-age and host-parasite-density. These host-features are thought to provide information on the not yet-understood complex relationships between host-immunity, host-genetics, parasite load, transmission burden and seasonality. Overall, the tendency has been to build monolithic predictive systems, despite malaria data being scanty and unprecise, that have been unable to generalize across malaria regions. These monolithic systems cannot fairly be tasked with predicting good local estimates of prevalence while at the same time being able to accurately detect extreme pattern-changes globally. On the contrary, our results show the feasibility of a data-driven locality-specialized malaria prevalence predicting system for a large metropolis of 3 million inhabitants in sub-Saharan West Africa. Our system can be used as a starting point to support the deployment of locally-relevant systems across malaria affected regions such as densely-populated metropolis of Lagos and Kano in Nigeria. At our tertiary level, LEMPS supports the readiness of our blood bank to sustain the near-zero mortality of our severe malarial anemia care pathways. At our primary community and peri-urban level LEMPS supports readiness for diagnosis and treatment of uncomplicated malaria. Our LEMPS could be fine-tuned to support regionally-dependent adaptability and readiness of healthcare pathways, each with their own critical bottlenecks, which is well recognised by the WHO as key for the global technical strategy for malaria. In rural settings, LEMPS could facilitate the use of locality-specific data to tackle their own critical bottlenecks as well as allowing the interaction with urban settings to achieve this. In this context, our LEMPS is a realisable step towards achieving truly data-driven digital global health.

More emphasis should be put on building meaningful multi-view data-driven locality-specific systems designed to harness local data for well-defined tasks such as the one presented in this study. Machine learning meta-models that take input from these regionally specialized systems could be most suited to provide vast regional epidemiological decision-support. We propose a deployment scenario where many regional centers, each a locally trained LEMPS node (harnessing such local data at its best), push their data and predictions into a distributed ledger that ensures consensus, consistency and immutability of information across participating nodes. New LEMPS nodes could use closer regional predictors while they gear up to produce their own refined predictors. As the network of locally specialized predictors grows, it opens the possibility of meta-learning and novelty detection algorithms to be applied for tasks such as early epidemic prediction. Our system provides a step towards supporting efficient distribution of resources that takes into account the different locality-specific characteristics of malaria affected regions. Equally important, such a distributed ledger should provide an interface by which global healthcare authorities, policy makers and malaria control programs interact and support their decisions with regionally relevant data.

Finally, our validated LEMPS system shows that local good-quality malaria longitudinal-data can be harnessed by current data-driven machine learning approaches to deliver locality-relevant predictions on burden of malaria. Reliable and adaptable malaria prediction systems can play key roles when deployed within a well-defined resource-stretched healthcare network as in the case of the large Ibadan metropolis where our system is deployed. In our settings, the system provides relevant next month prevalence estimates that are used for aiding decision making on critical aspects of urban to peri-urban care pathways. The deployment simplicity of our LEMPS provides an incentive for other sub-Saharan centers, by enabling decision-support using their own routinely collected malaria data, to consider sustainable digital global health approaches to tackle challenges on healthcare provision in the region.

**Financial Disclosure**

This work was supported by the College of Medicine of the University of Ibadan, Ibadan, Nigeria (comui.edu.ng) (BJB, OS, DF-R); the United Kingdom Medical Research Council (mrc.ukri.org) (Grant Number: MC_U117585869) (DF-R); the United Kingdom Engineering and Physical Sciences Research Council (epsrc.ukri.org) (Grant Number: EP/P028608/1) (DF-R) and; the Department of Computer Science (cs.ucl.ac.uk), Faculty of Engineering Sciences, University College London (ucl.ac.uk), United Kingdom (DF-R). The Childhood Malaria Research Group (CMRG) is a joint malaria research and innovation equal-partnership between the College of Medicine of University of Ibadan, Nigeria and University College London, London, UK. The funders had no role in study design, data collection and analysis, decision to publish, or preparation of the manuscript.

**Competing Interests**

The authors have declared that no competing interests exist.

**Data Availability Statement**

LEMPS data and code are available upon request to the Childhood Malaria Research Group (CMRG), Department of Paediatrics, College of Medicine of University of Ibadan, University College Hospital, Ibadan, Nigeria. Emails: CMRG-Nigeria (paedcomui@yahoo.com); CMRG-UK (delmiro.fernandez-reyes@ucl.ac.uk) and will be also made accessible via COMUI and UCL web-sites.

**Author Contributions**

DF-R, BJB and OS designed the study. OS, BJB, DF-R, AEO, WAA, FA, OK, KO, FOA, SO, WAS, IL carried out the study. OS, BJB, DF-R, AEO, WAA, FA, OK, KO, FOA, SO, WAS, IL, GO, ME, NKA carried patient recruitment at the CMRG, Ibadan, Nigeria. DF-R, BJB, ME, GO, MS, VP, PM, RC, JS-T, MAS, BJB, IL, OS, GO (the FAST-Mal team) collected and processed 2015 to 2017 validation datasets. DF-R, BJB, GO, ME, AP, PM processed datasets. DF-R, AP, PM, FC designed and coded the system and carried out computational work. DF-R, AP, PM, FC analyzed the data. DF-R, BJB, AP, PM, FC wrote the manuscript. DF-R is project lead and senior corresponding author.

**Acknowledgements**

The authors thank all study participants. We thank consultants, clinical registrars, nurses and clinical laboratory staff at the College of Medicine of the University of Ibadan, Nigeria for all the support they provided for the present study. We thank the research and administrative staff who provided support at the College of Medicine of the University of Ibadan, Nigeria and at the Faculty of Engineering, University College London, United Kingdom.

**Materials & Correspondence**

Delmiro Fernandez-Reyes, MD, MSc, DPhil.

Address 1: University College London, Department of Computer Science, Faculty of Engineering, Gower Street, London, WC1E 6BT, United Kingdom.

Address 2: College of Medicine of the University of Ibadan, Department of Paediatrics, University College Hospital, Ibadan, Nigeria.

Phone UK & Worldwide: +44(0)7956661869

email: Delmiro.Fernandez-Reyes@ucl.ac.uk

**Table 1.** Clinical characteristics, aggregated by year, of training and validation cohorts.

|  | Year | Screened N | F / M (%) | MP (+ve) N | MP (-ve) N | Age MP (+ve) months median (IQR) | Age MP (-ve) months median (IQR) | MP Density MPs/µl log[mean (se)] |
|---|---|---|---|---|---|---|---|---|
| **TRAINING (DTRAS)** | 1996 | 2,295 | 47 / 53 | 627 | 1,668 | 36 (42) | 24 (50) | 4.1 (3.0) |
| | 1997 | 3,732 | 44 / 56 | 879 | 2,853 | 36 (55) | 24 (48) | 4.2 (2.8) |
| | 1998 | 2,705 | 42 / 58 | 1,300 | 1,405 | 36 (44) | 27 (50) | 4.2 (3.0) |
| | 1999 | 2,380 | 45 / 55 | 919 | 1,461 | 30 (43) | 29 (50) | 4.2 (3.0) |
| | 2000 | 2,404 | 41 / 59 | 669 | 1,735 | 36 (43) | 24 (45) | 4.3 (3.2) |
| | 2001 | 2,656 | 49 / 51 | 762 | 1,894 | 51 (58) | 50 (64) | 4.5 (3.2) |
| | 2002 | 2,688 | 48 / 52 | 872 | 1,816 | 48 (55) | 48 (60) | 4.4 (3.1) |
| | 2003 | 2,716 | 45 / 55 | 915 | 1,801 | 36 (43) | 34 (52) | 4.3 (3.0) |
| | 2004 | 4,123 | 45 / 55 | 856 | 3,267 | 36 (42) | 24 (52) | 3.8 (2.9) |
| | 2005 | 5,504 | 44 / 56 | 1,284 | 4,220 | 36 (45) | 24 (49) | 4.1 (2.9) |
| | 2006 | 5,497 | 43 / 57 | 1,055 | 4,442 | 30 (45) | 26 (48) | 4.2 (3.0) |
| | 2007 | 6,882 | 42 / 58 | 1,083 | 5,799 | 31 (43) | 24 (48) | 4.2 (2.9) |
| | 2008 | 5,341 | 42 / 58 | 966 | 4,375 | 36 (42) | 27 (49) | 4.2 (2.9) |
| | 2009 | 6,585 | 43 / 57 | 838 | 5,747 | 42 (54) | 28 (48) | 4.4 (3.1) |
| | 2010 | 6,258 | 43 / 57 | 965 | 5,293 | 36 (43) | 21 (49) | 4.5 (3.2) |
| | 2011 | 5,207 | 43 / 57 | 689 | 4,518 | 35 (54) | 28 (44) | 4.4 (3.2) |
| | 2012 | 5,180 | 42 / 58 | 737 | 4,443 | 39 (54) | 22 (51) | 4.5 (3.3) |
| | 2013 | 5,014 | 44 / 56 | 589 | 4,425 | 36 (45) | 23 (49) | 4.6 (3.4) |
| | 2014 | 5,073 | 43 / 57 | 898 | 4,175 | 30 (50) | 24 (48) | 4.6 (3.3) |
| | *TOTAL* | *82,240* | *44 / 56* | *16,903* | *65,337* | *36 (47)* | *28 (50)* | *4.3 (3.0)* |
| **VALIDATION (DVASL)** | 2015 | 3,091 | 49 / 51 | 551 | 2,540 | 40 (55) | 32 (49) | 4.6 (3.2) |
| | 2016 | 3,743 | 47 / 53 | 534 | 3,209 | 41 (48) | 34 (51) | 4.5 (3.1) |
| | 2017 | 3,737 | 48 / 52 | 641 | 3,096 | 48 (65) | 68 (62) | 4.5 (2.5) |
| | *TOTAL* | *10,571* | *48 / 52* | *1,726* | *8,845* | *43 (56)* | *45 (54)* | *4.5 (2.9)* |

IQR= interquartile range. MP Density = malaria parasites per microliter.

**Table 2.** Overall characteristics of Training (DTRAS) and Validation (DVALS) of Ibadan Dataset

| Ibadan Dataset | Number Years | Dates | Months | *M* row-vector instances | *N* variables |
|---|---|---|---|---|---|
| Training Set (DTRAS) | 19 | 1996 - 2014 | Jan-Dec | 228 | 15 |
| Validation Set (DVALS) | 3 | 2015 - 2017 | Jan-Dec | 36 | 15 |

**Table 3.** Ibadan dataset monthly aggregated variables

| | Instance Variables (*N*=15) | | | |
|---|---|---|---|---|
| **Index *x*** | ***Variable Name*** | **Description** | **Units** | **Aggregate** |
| 1 | *year* | Year $i$ | $i$ = 1996 to 2017 | No |
| 2 | *month* | Month $j$ | $j$ = 1 to 12 | No |
| 3 | *number-screened* | Total number screened | Integer | Sum $Month_{i,j}$ |
| 4 | *median-age-neg* | Median age of malaria-negative | Age (months) | Median $Month_{i,j}$ |
| 5 | *median-age-pos* | Median age of malaria-positive | Age (months) | Median $Month_{i,j}$ |
| 6 | *iqr-age-neg* | IQR age malaria-negative | Age (months) | IQR $Month_{i,j}$ |
| 7 | *iqr-age-pos* | IQR age malaria-positive | Age (months) | IQR $Month_{i,j}$ |
| 8 | *x-pd* | Mean of blood parasite densities* | MPs/μl | Mean $Month_{i,j}$ |
| 9 | *sd-pd* | STD of blood parasite densities* | MPs/μl | STD $Month_{i,j}$ |
| 10 | *mm-rf* | Month total rainfall | mm | Sum $Month_{i,j}$ |
| 11 | *mmP-rf* | Proportion of that Year $i$ total rainfall | Proportion | |
| 12 | *min-temp* | $Month_{i,j}$ minimum temperature | Celsius | |
| 13 | *max-temp* | $Month_{i,j}$ maximum temperature | Celsius | |
| 14 | *x-temp* | Month Mean temperature | Celsius | Mean $Month_{i,j}$ |
| 15 | *prep* | $Month_{i,j}$ malaria prevalence** | Proportion | |

| | | | | | Each row-vector d of D (index *x*)*** | | | | | | | | | |
|---|---|---|---|---|---|---|---|---|---|---|---|---|---|---|
| 1 | 2 | 3 | 4 | 5 | 6 | 7 | 8 | 9 | 10 | 11 | 12 | 13 | 14 | 15 |

*Parasite Density (pd) = malaria parasites per microliter (MPs/μl)
= (number-observed-malaria-parasites/number-observed White Blood Cells (WBC)) * 8000
**Proportion of screened with confirmed malaria.
***row-vector *d* of *D* (variable 1)-(variable 2) form the unique year-month key for that instance.
IQR = Inter-Quartile-Range. STD = Standard Deviation.

**Table 4.** Encoding of Regression Tasks

| **Dataset Task+1 M$_m$-DTRAS** & **Dataset Task+1 M$_m$-DVALS** |
|---|
| (DT1M$_m$-DTRAS) & (DT1M$_m$-DVALS) |
| where *m* = 1 to 6; |
| **DTRAS** *i*=year (1996 to 2014); **DVALS** *i*=year (2015 t0 2017); *j*=month (Jan to Dec) |
| Each row-vector **d** of DT1M$_m$-DTRAS = $d_{m,DTRASi,j} = DTRAS(x_{m,DTRASi,j}, y_{DTRASi,j})$ |
| Each row-vector **d** of DT1M$_m$-DVALS = $d_{m,DVALSi,j} = DVALS(x_{m,DVALSi,j}, y_{DVALSi,j})$ |
| Target $y_{DTRASi,j}$ = malaria prevalence of **DTRAS i, j** following month |
| Target $y_{DVALSi,j}$ = malaria prevalence of **DVALS i, j** following month |
| $x_{m,i,j}$ = append *i, j* previous (*m*-1) months = [1 by *N*=(15 x *m*) row-vector] |

| DTask+1 M$_m$ | M6 | M5 | M4 | M3 | M2 | M1 | d$_{all}$ | d$_{final}$ remove year column(s) |
|---|---|---|---|---|---|---|---|---|
| Task+1 M$_m$=1 (DT1M1) | NO | NO | NO | NO | NO | YES | 1 by 16 | 1 by 15 |
| Task+1 M$_m$=2 (DT1M2) | NO | NO | NO | NO | YES | YES | 1 by 31 | 1 by 29 |
| Task+1 M$_m$=3 (DT1M3) | NO | NO | NO | YES | YES | YES | 1 by 46 | 1 by 43 |
| Task+1 M$_m$=4 (DT1M4) | NO | NO | YES | YES | YES | YES | 1 by 61 | 1 by 57 |
| Task+1 M$_m$=5 (DT1M5) | NO | YES | YES | YES | YES | YES | 1 by 76 | 1 by 71 |
| Task+1 M$_m$=6 (DT1M6) | YES | YES | YES | YES | YES | YES | 1 by 91 | 1 by 85 |

Example of an encoded row-vector *see Table 3 for variable index and names

| **Example of (index *x*)* features of a 1 by 15 d$_{final}$ row-vector of DT1M1 Tasks** | | | | | | | | | | | | | | | |
|---|---|---|---|---|---|---|---|---|---|---|---|---|---|---|---|
| 2 | 3 | 4 | 5 | 6 | 7 | 8 | 9 | 10 | 11 | 12 | 13 | 14 | 15 | *y* |
| features | | | | | | | | | | | | | | target |

**FIGURE LEGENDS**

**Figure 1.** Study site geolocation and its monthly burden of malaria from 1996 to 2017.

**a.** *Left and Centre*: geographical location of the third largest urban large densely-populated setting in Nigeria, the City of Ibadan. *Right*: Ibadan's urban boundary; dropped-pin shows location of UCH Ibadan; red-balls shows location of primary and community centers.

**b.** Ibadan dataset 3D surface-plot showing monthly mean malaria prevalence (y-axis and heat map); month (x-axis); year (z-axis) from 1996 to 2017.

**Figure 2.** Machine learning algorithms parametrization, evaluation and model selection on the Ibadan training DTRAS dataset.

DTRAS= Ibadan Dataset Training Set [from 1996 to 2014]; EN=Elastic Net; LASSO=Least Absolute Shrinkage and Selection Operator; RR=Ridge Regression; LARS=Least Angle Regression; AIC= Akaike Information Criterion; BIC=Bayesian Information Criterion; SVR= Support Vector Regression; $a$ = regularization strength parameter; C=SVR margin parameter; $g$=SVR sigma gaussian-kernel parameter; MAE= Mean Absolute Error; MSE= Mean Square Error; X=features; y=true prevalence; $\hat{y}$=predicted prevalence; [1](using 5-fold cross validation); [2](L1Ratio=0.5).

**Figure 3.** MAE and MSE errors of used machine learning approaches on training DTRAS dataset

**a.** Mean and Standard Deviation MAE.

**b.** Mean and Standard Deviation MSE.

Algorithms in order from left to right per each regression task DT1M1 to DT1M2: EN (filled circles); LASSO (filled squares); RR (filled up-triangles); LASSO-LARS (filled down-triangles); LASSO-LARS-AIC (empty circles); LASSO-LARS-BIC (empty squares); RF (empty up-triangles) and SVR (empty down-triangles). DTRAS= Ibadan Dataset Training Set [from 1996 to 2014]; EN=Elastic Net; LASSO=Least Absolute Shrinkage and Selection Operator; RR=Ridge Regression; LARS=Least Angle Regression; AIC= Akaike Information Criterion; BIC=Bayesian Information Criterion; SVR= Support Vector Regression; MAE= Mean Absolute Error; MSE= Mean Square Error.

**Figure 4.** The Locality-specific Elastic Net based Malaria Prevalence prediction System (LEMPS)

**a.** LEMPS regularization strength and L1-norm ratio model selection on training DTRAS dataset.

**b.** LEMPS validation on DVALS dataset.

DTRAS=Ibadan Dataset Training Set [from 1996 to 2014]; DVALS=Ibadan Dataset Validation Set [from 2015 to 2017]; $a$ = regularization strength parameter; MAE= Mean Absolute Error; MSE= Mean Square Error; X=features; y=true prevalence; $\hat{y}$=predicted prevalence; [1](using 5-fold cross validation);

**Figure 5.** LEMPS performance and best parameters range on training DTRAS dataset

**a.** Mean and Standard Deviation MAE.

**b.** Mean and Standard Deviation MSE.

**c.** Mean and Standard Deviation of regularization strength parameter $a$.

**d.** Median and Interquartile Range of L1/L2 norm ratio parameter L1Ratio

DTRAS=Ibadan Dataset Training Set [from 1996 to 2014]; MAE= Mean Absolute Error; MSE= Mean Square Error; pre=prevalence.

**Figure 6.** LEMPS evaluation of true *vs.* predicted prevalence values on Held-Out Test Set over $10^3$ random samplings of the training DTRAS dataset.

**a.** to **f.** True prevalence value $y$ (dotted black line) vs. mean predicted prevalence value $\hat{y}$ (red line) for all the regression tasks DT1M1 to DT1M6 respectively over $10^3$ random samplings. DTRAS=Ibadan Dataset Training Set [from 1996 to 2014].

**Figure 7.** LEMPS performance on validation set DVALS.

Final LEMPS system yearly MAE, MSE and PCC on 2015 (filled orange circles), 2016 (filled orange squares) and 2017 (filled orange triangles) DVALS validation set on all regression tasks DT1M1 to DT1M6. DVALS= Ibadan Dataset Validation Set [from 2015 to 2017]; MAE= Mean Absolute Error; MSE= Mean Square Error; PCC=Pearson Correlation Coefficient; pre=prevalence.

**Figure 8.** Scatter 2D plots of LEMPS true and predicted prevalence on validation set DVALS.

For all validation years 2015, 2016, 2017 and all regression tasks DT1M1 to DT1M6. *x-axis*: true prevalence value $y$; *y-axis*: EN predicted prevalence value $\hat{y}$; red dots= dry season; blue dots= rainy season. DVALS=Ibadan Dataset Validation Set [from 2015 to 2017].

**Figure 9.** LEMPS predicted prevalence on validation set within locality relevant tolerance-error.

LEMPS predicted prevalence for all validation years 2015, 2016, 2017 and all regression tasks DT1M1 to DT1M6 (orange, blue, red, purple, green, yellow filled squares respectively) plotted against the true prevalence value (black circles) and true prevalence value +0.1 to -0.05 tolerance-error (shaded grey area).

**Figure 10.** Proposed LEMPS Deployment Scenarios

**a.** *Scenario 1*: a locally deployed LEMPS is parametrized to provide accurate regional-specific estimates of next-month malaria prevalence. *Scenario 2*: a network of locally deployed LEMPS systems push their regional predictions to a distributed ledger that ensures consistency, immutability and trust among participating nodes and provides access to relevant stakeholders.

**b.** A high-level view of the interaction of individual local LEMPS where a distributed ledger provides an interface for third parties to access the fine-grained burden of disease data.

**REFERENCES**

1        World Health Organization. World Malaria Report 2016. Geneva: World Health Organization, 2016 DOI:CC BY-NC-SA 3.0 IGO.

2        World Health Organisation. Malaria in children under five. 2018. http://www.who.int/malaria/areas/high_risk_groups/children/en/.

3        World Health Organization. Global Technical Strategy for Malaria 2016-2030. Resolution WHA68.2. 2015 http://www.who.int/malaria/areas/global_technical_strategy/en/.

4        World Health Organization. Overview of malaria surveillance. 2015. http://www.who.int/malaria/areas/surveillance/overview/en/.

5        World Health Organization. WHO Nigeria country profile. 2016. http://www.who.int/malaria/publications/country-profiles/profile_nga_en.pdf.

6        Zinszer K, Verma AD, Charland K, *et al.* A scoping review of malaria forecasting: past work and future directions. *BMJ Open* 2012; **2**. DOI:10.1136/bmjopen-2012-001992.

7        Kouwaye B, Rossi F, Fonton N, *et al.* Predicting local malaria exposure using a Lasso-based two-level cross validation algorithm. *PLoS One* 2017. DOI:10.1371/journal.pone.0187234.

8        Zacarias OP, Boström H. Predicting the Incidence of Malaria Cases in Mozambique Using Regression Trees and Forests. *Int J Comput Sci Electron Eng* 2013.

9        Sharma V, Kumar A, Panat L, Karajkhede G, Lele A. Malaria Outbreak Prediction Model Using Machine Learning. *Int J Adv Res Comput Eng Technol* 2015.

10       Chintalapati S, Sohani SK, Kumar D, *et al.* A Support Vector Machine-Firefly Algorithm based forecasting model to determine malaria transmission. *Neurocomputing* 2014; **129**: 279–288.

11  Modu B, Polovina N, Lan Y, Konur S, Asyhari A, Peng Y. Towards a Predictive Analytics-Based Intelligent Malaria Outbreak Warning System. *Appl Sci* 2017; **7**: 836.

12  Burté F, Brown BJ, Orimadegun AE, *et al.* Severe Childhood Malaria Syndromes Defined by Plasma Proteome Profiles. *PLoS One* 2012. DOI:10.1371/journal.pone.0049778.

13  Burté F, Brown BJ, Orimadegun AE, *et al.* Circulatory hepcidin is associated with the anti-inflammatory response but not with iron or anemic status in childhood malaria. *Blood* 2013. DOI:10.1182/blood-2012-10-461418.

14  Ajetunmobi WA, Orimadegun AE, Brown BJ, *et al.* Haemoglobinuria among children with severe malaria attending tertiary care in Ibadan, Nigeria. *Malar J* 2012; **11**: 336.

15  Bachmann J, Burté F, Pramana S, *et al.* Affinity Proteomics Reveals Elevated Muscle Proteins in Plasma of Children with Cerebral Malaria. *PLoS Pathog* 2014; **10**. DOI:10.1371/journal.ppat.1004038.

16  Marquet S, Conte I, Poudiougou B, *et al.* The IL17F and IL17RA Genetic Variants Increase Risk of Cerebral Malaria in Two African Populations. *Infect Immun* 2016; **82**: 590–7.

17  Safeukui I, Gomez ND, Adelani AA, *et al.* Malaria induces anemia through CD8+T Cell-dependent parasite clearance and erythrocyte removal in the spleen. *MBio* 2015; **6**. DOI:10.1128/mBio.02493-14.

18  Marquet S, Conte I, Poudiougou B, *et al.* A functional IL22 polymorphism (rs2227473) is associated with predisposition to childhood cerebral malaria. *Sci Rep* 2017; **7**. DOI:10.1038/srep41636.

19  Abah SE, Burté F, Marquet S, *et al.* Low plasma haptoglobin is a risk factor for life-threatening childhood severe malarial anemia and not an exclusive consequence of hemolysis. *Sci Rep* 2018. DOI:10.1038/s41598-018-35944-w.

20      Fernandez-Reyes D, Shawe-Taylor J, Srinivasan MA, Brown BJ. Fast Accurate Scalable Malaria Diagnosis. https://uclfastmal.wordpress.com.

21      World Health Organization. Malaria Parasite Counting: Standard Operating Procedure MM-SOP-09. 2016.

22      Efron B, Hastie T, Johnstone I, *et al.* Least angle regression. *Ann Stat* 2004; **32**: 407–99.

23      Breiman L. Random forests. *Mach Learn* 2001. DOI:10.1023/A:1010933404324.

24      Cristianini N, Shawe-Taylor J. An introduction to Support Vector Machines. 2000 DOI:10.1017/S0263574700232827.

25      Drucker H, Burges C, Kaufman L, Smola A, Vapnik V. Support Vector Regression Machines. *Adv Neural Inf Process Syst* 1996; **9**: 155–61.

26      Akaike H. A New Look at the Statistical Model Identification. *IEEE Trans Automat Contr* 1974; **19**: 716–23.

27      Schwarz G. Estimating the Dimension of a Model. *Ann Stat* 1978; **5**: 461–4.

28      Pedregosa F, Varoquaux G, Gramfort A, *et al.* Scikit-learn: Machine Learning in Python. *J Mach Learn Res* 2011; **12**: 2825–30.

29      Pedregosa F, Varoquaux G, Gramfort A, *et al.* Scikit-learn: Machine Learning in Python. *J Mach Learn Res* 2011; **12**: 2825–30.

**Figure 1.**

a.

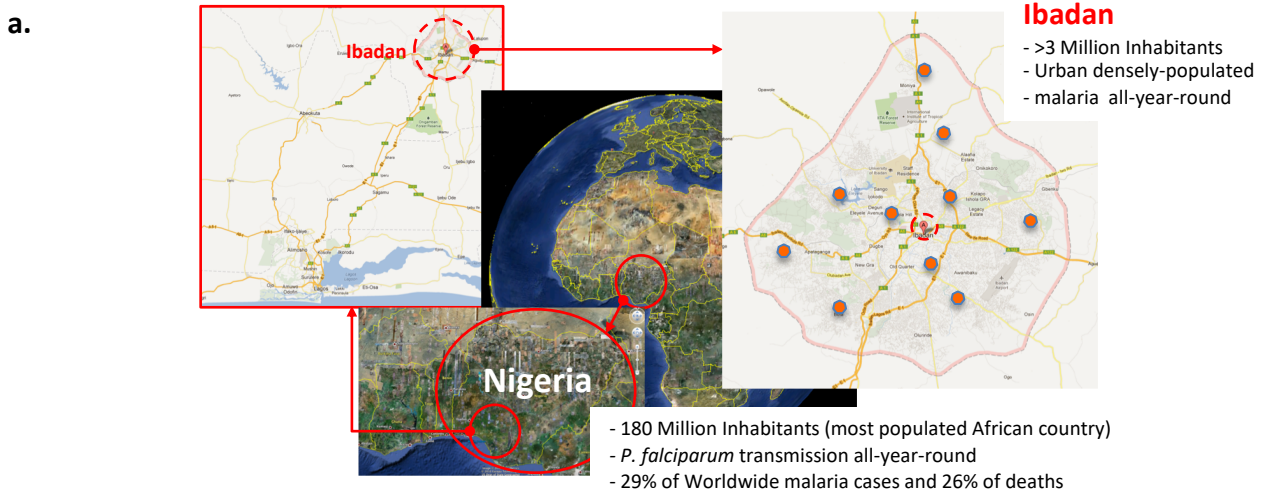

b.

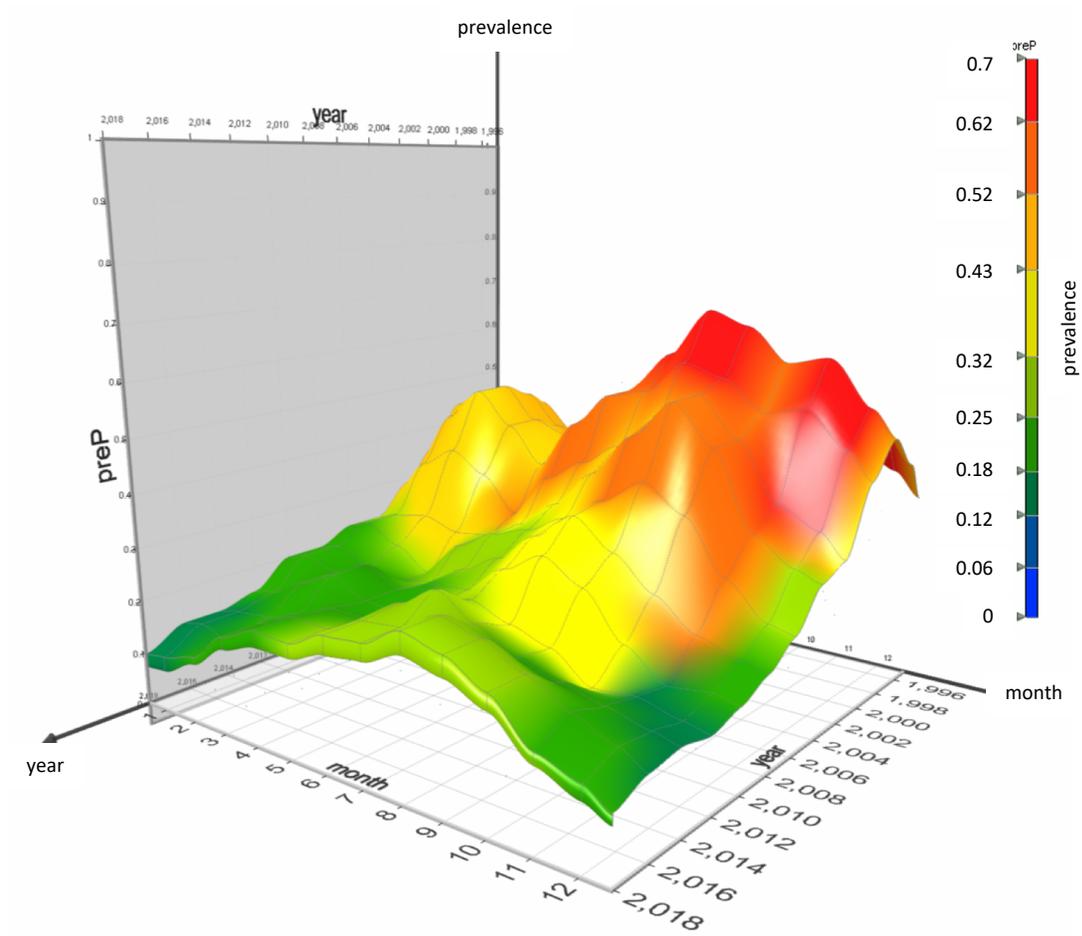

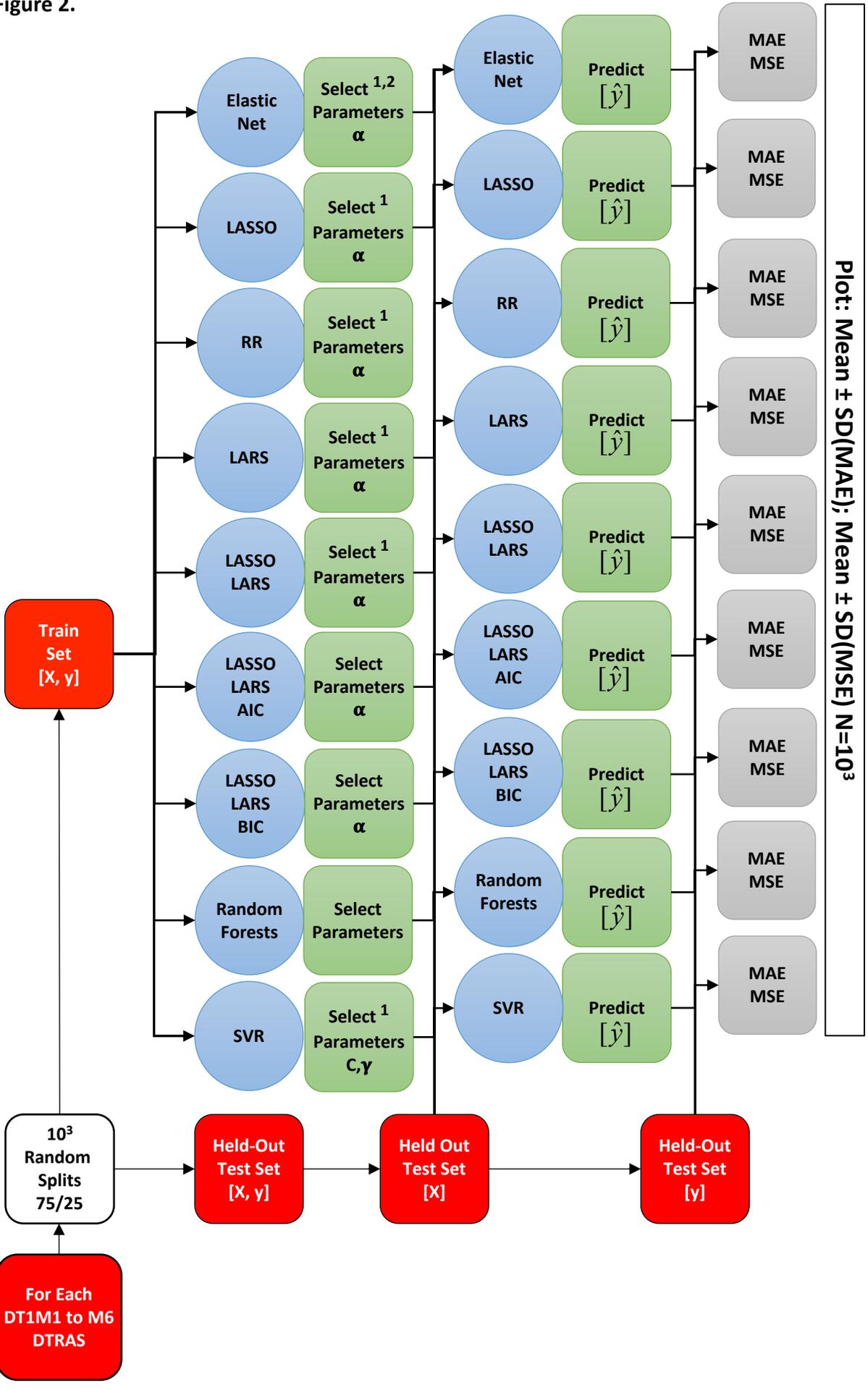

Figure 2.

**Figure 3.**

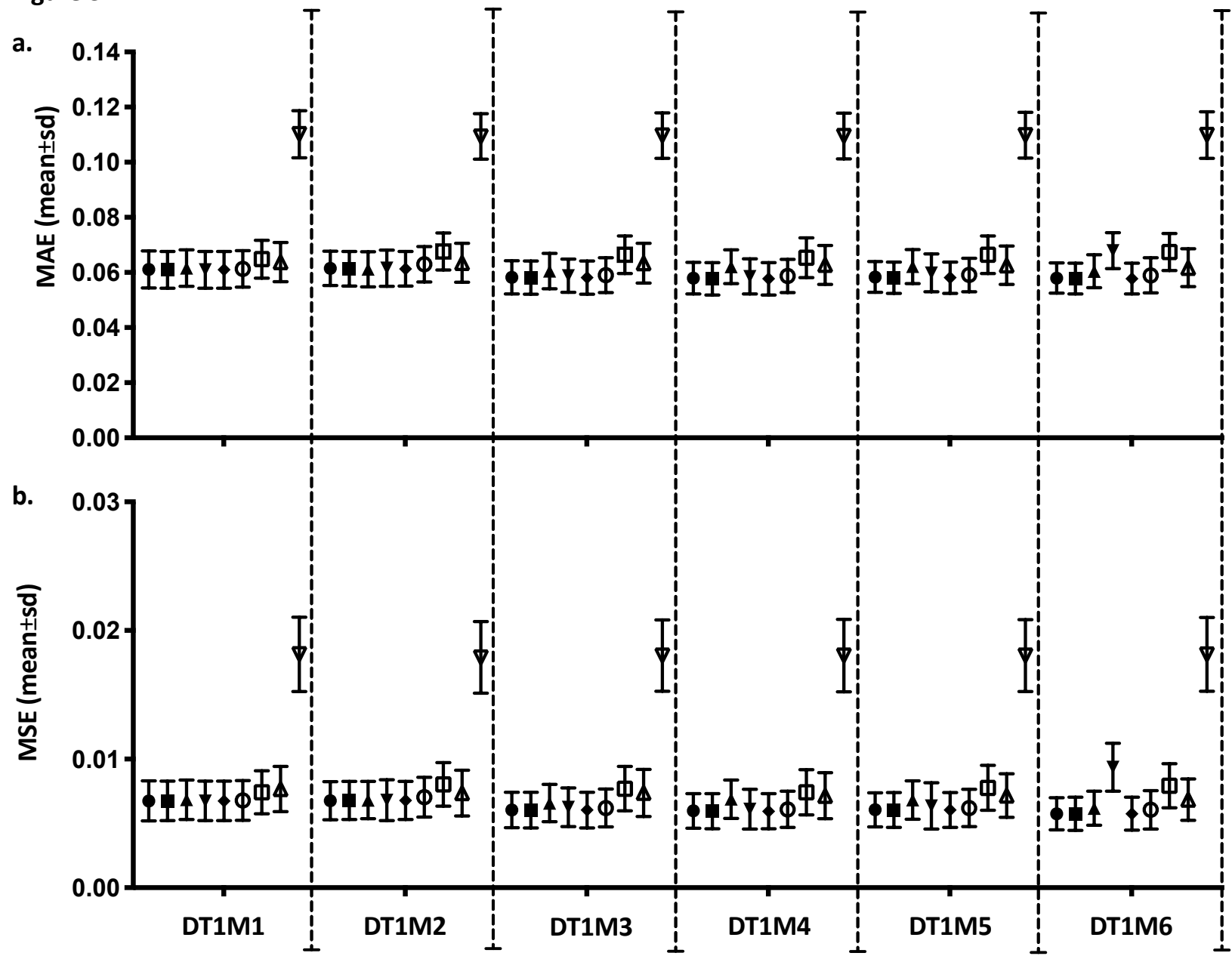

**Figure 4.**

**a. EN alpha and L1Ratio Selection**

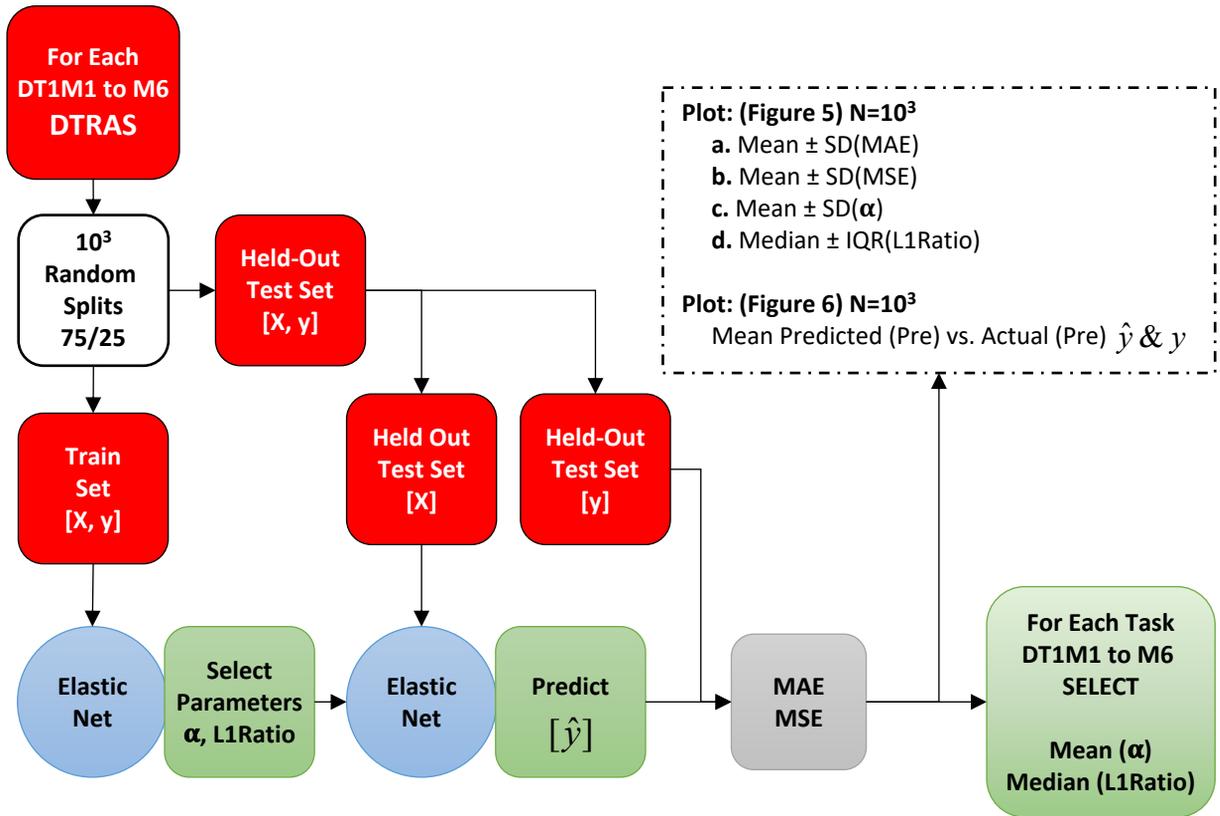

**b. Validation**

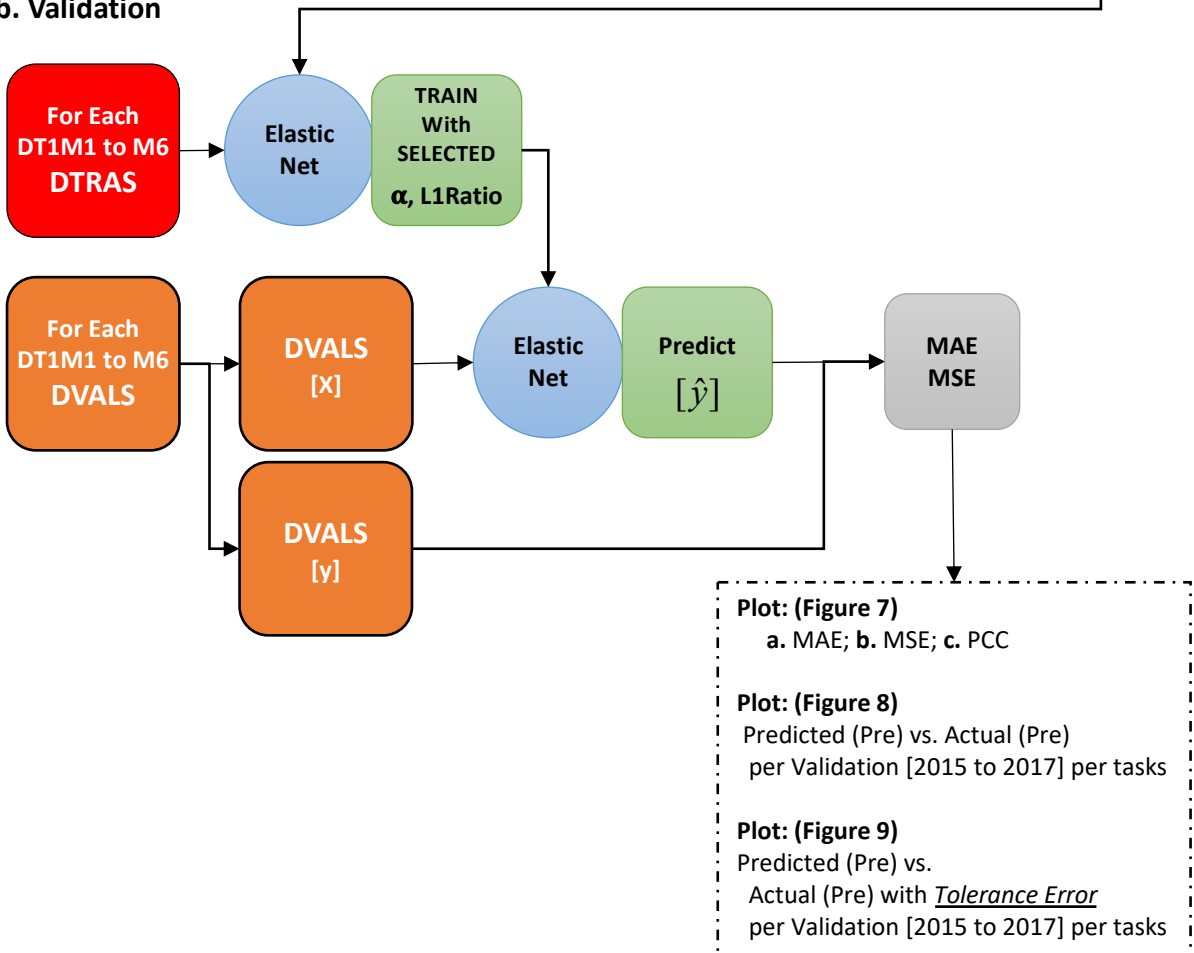

**Figure 5.**

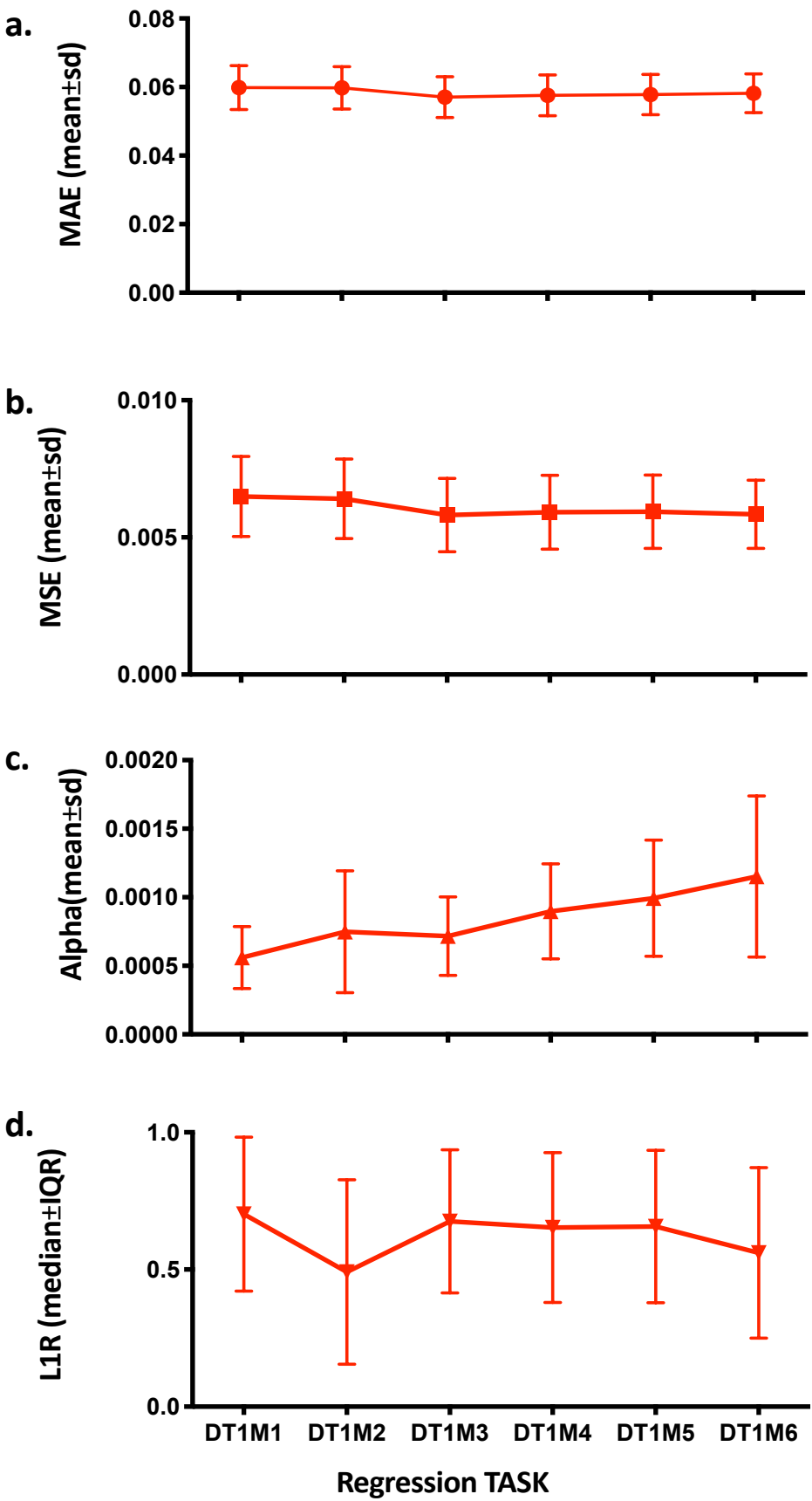

**Figure 6.**

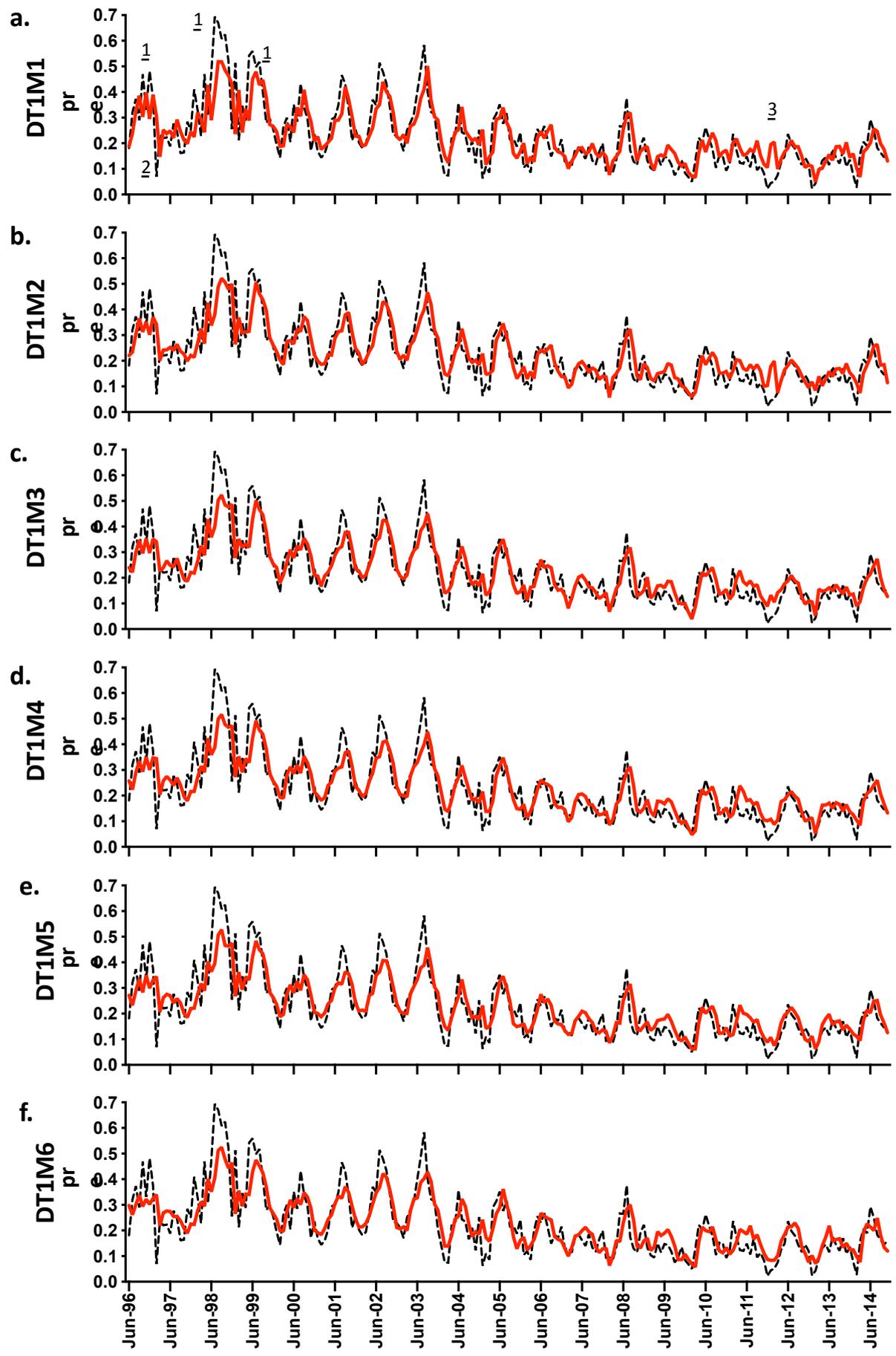

**Figure 7.**

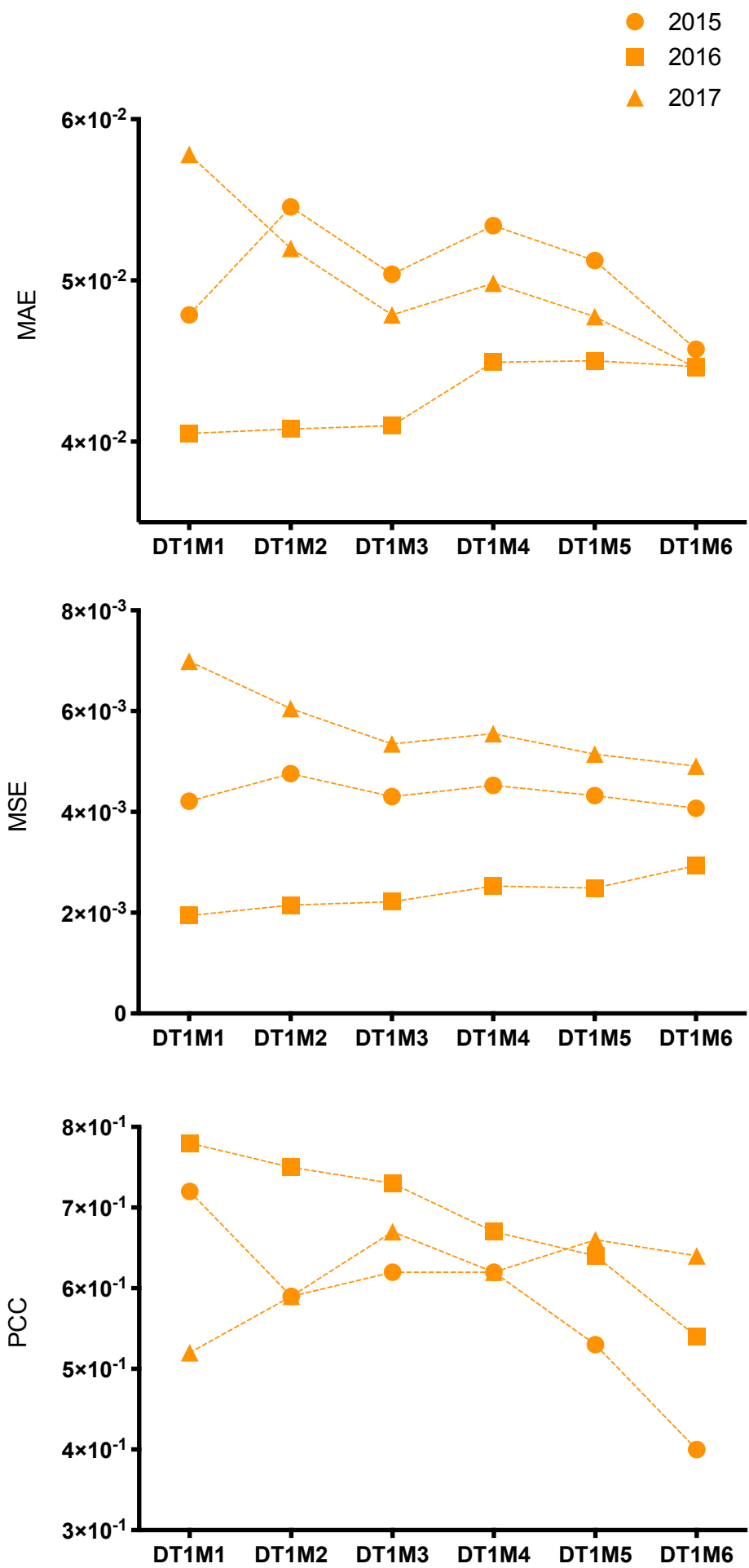

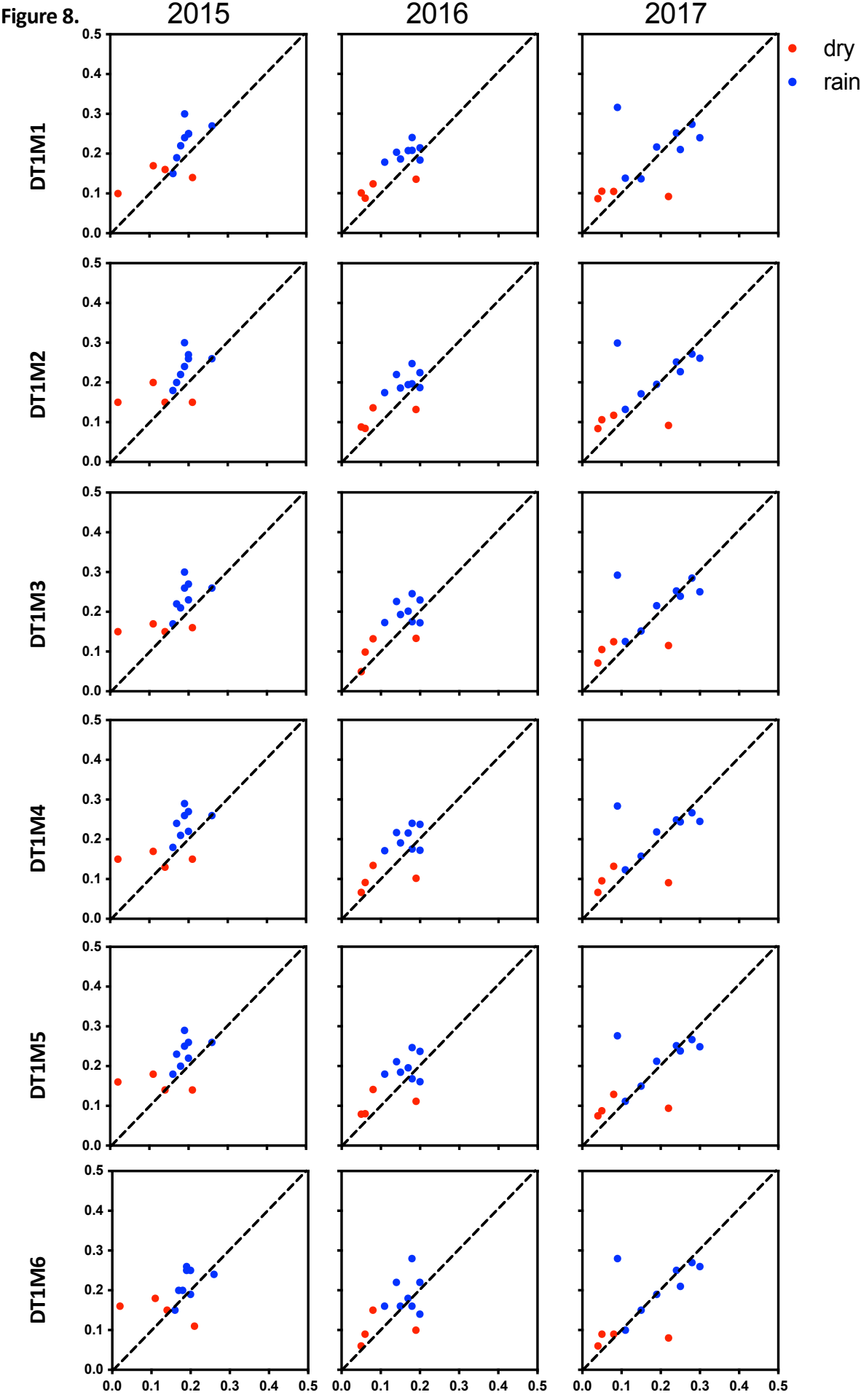

Figure 8.

**Figure 9.**

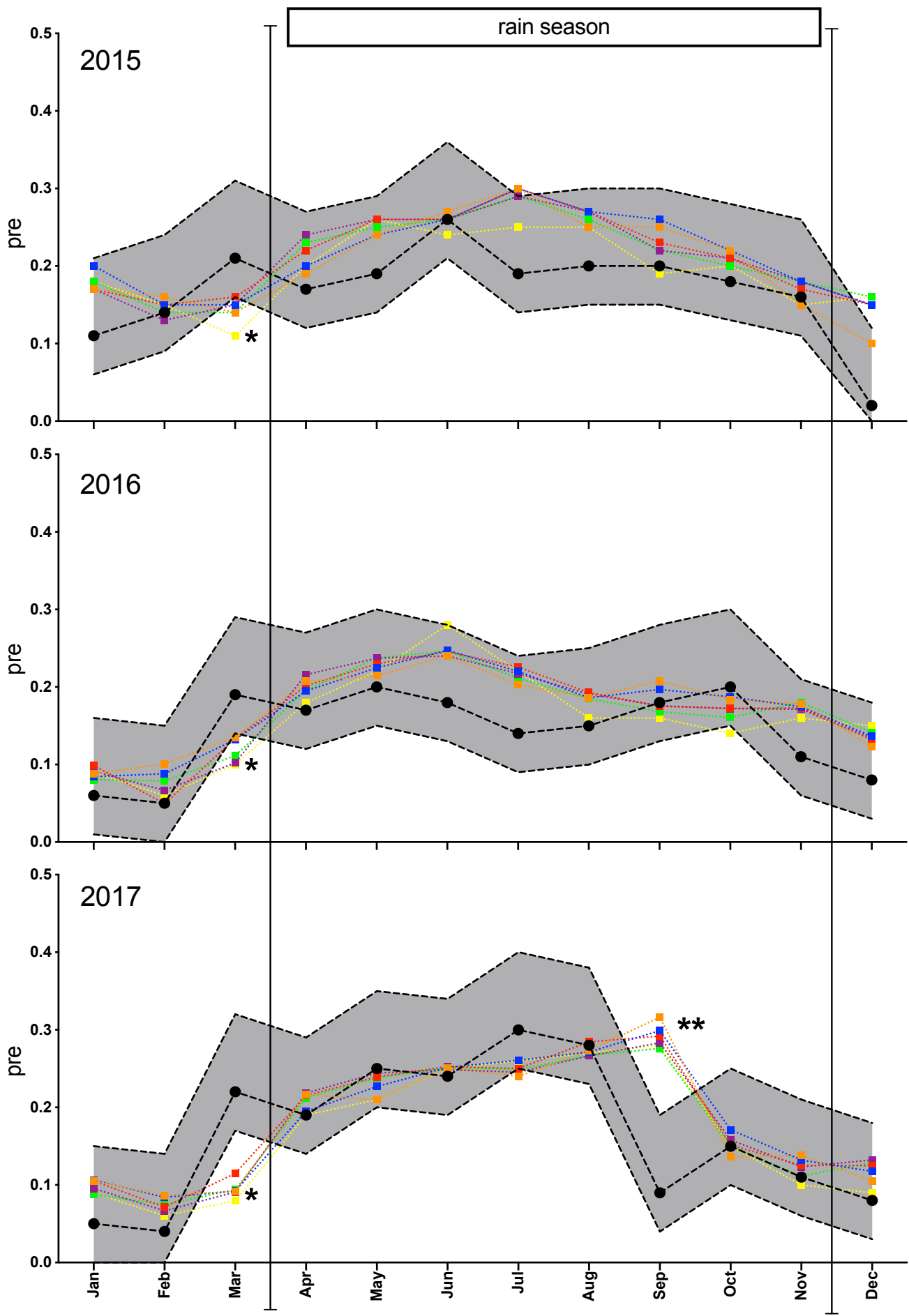

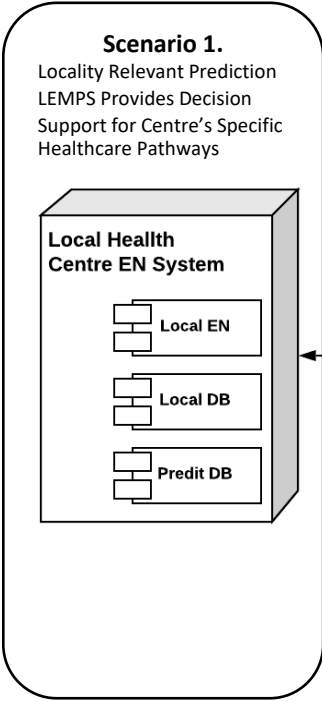
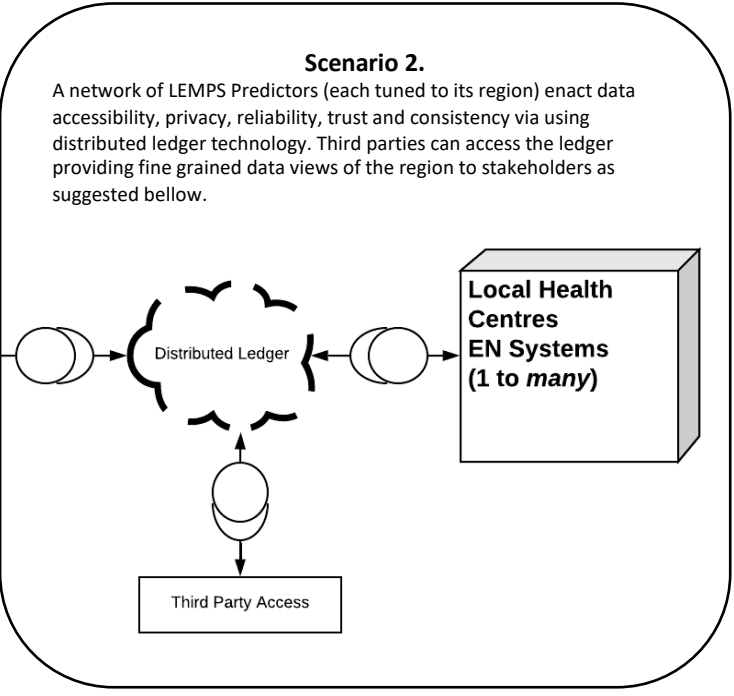
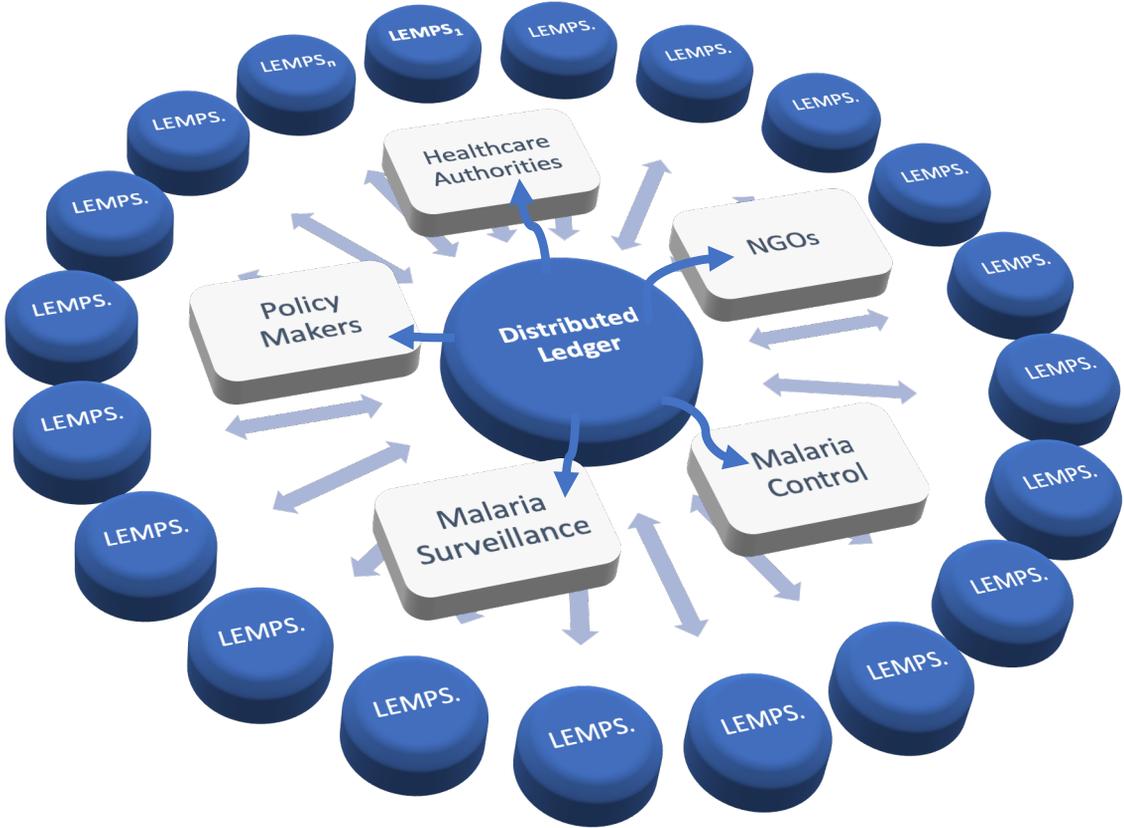